\documentclass[ba, preprint]{imsart}
\pubyear{2021}
\volume{TBA}
\issue{TBA}
\firstpage{1}
\lastpage{1}

\usepackage{lipsum}
\usepackage{bm}
\usepackage{dsfont}
\usepackage{amsthm}
\usepackage{amsmath}
\usepackage{amssymb}
\usepackage{amsfonts}
\usepackage{natbib}
\setcitestyle{notesep={ }}
\usepackage[colorlinks,citecolor=blue,urlcolor=blue,filecolor=blue,backref=page]{hyperref}
\usepackage{graphicx}
\usepackage[utf8]{inputenc}
\usepackage[T1]{fontenc}
\usepackage{booktabs}

\usepackage{algorithm}
\usepackage[noend]{algpseudocode}
\algnewcommand\algorithmicinput{\textbf{Input:}}
\algnewcommand\algorithmicoutput{\textbf{Output:}}
\algnewcommand\Input{\item[\algorithmicinput]}%
\algnewcommand\Output{\item[\algorithmicoutput]}%

\DeclareMathOperator*{\argmax}{arg\,max}

\newcommand\smallO{
  \mathchoice
    {{\scriptstyle\mathcal{O}}}
    {{\scriptstyle\mathcal{O}}}
    {{\scriptscriptstyle\mathcal{O}}}
    {\scalebox{.7}{$\scriptscriptstyle\mathcal{O}$}}
  }

\startlocaldefs
\numberwithin{equation}{section}
\theoremstyle{plain}

\endlocaldefs

\usepackage{xcolor}

\usepackage{siunitx}
\usepackage{multirow}
\usepackage{booktabs}
\usepackage{etoolbox}
\newcommand{\B}{\fontseries{b}\selectfont}

\newcommand{\vint}{v}  
\newcommand{\vintbf}{\bm{\vint}}
\newcommand{\thetab}{\bm{\theta}}
\newcommand{\dbf}{\mathbf{d}}
\newcommand{\dbfopt}{\mathbf{d}^\ast}
\newcommand{\ybf}{\mathbf{y}}

\newcommand{\NN}{T_{\bm{\psi}}(\bm{\vint}, \ybf)}
\newcommand{\NNopt}{T^{\ast}_{\bm{\psi}}(\bm{\vint}, \ybf)}
\newcommand{\Ej}{\mathbb{E}_{p(\bm{\vint}, \ybf \mid \dbf)}}
\newcommand{\Em}{\mathbb{E}_{p(\bm{\vint}) p(\ybf \mid \dbf)}}

\newcommand{\psib}{\bm{\psi}}

\newcommand{\likeint}{p(\ybf|\bm{\vint},\dbf)}
\newcommand{\postint}{p(\bm{\vint}|\ybf,\dbf)}
\newcommand{\jointint}{p(\bm{\vint},\ybf|\dbf)}
\newcommand{\margint}{p(\ybf|\dbf)}
\newcommand{\priorint}{p(\bm{\vint})}

\newcommand{\like}{p(\ybf|\thetab,\dbf)}
\newcommand{\post}{p(\thetab|\ybf,\dbf)}

\newcommand{\prior}{p(\thetab)}

\newcommand{\likemd}{p(\ybf|m,\dbf)}
\newcommand{\postmd}{p(m|\ybf,\dbf)}

\newcommand{\priormd}{p(m)}

\newcommand{\likemdpe}{p(\ybf|\thetab_m,m,\dbf)}
\newcommand{\postmdpe}{p(\thetab_m,m|\ybf,\dbf)}

\newcommand{\priormdpe}{p(\thetab_m,m)}

\newcommand{\likefp}{p(\ybf_T|\ybf,\dbf, \dbf_T)}

\newcommand{\margfp}{p(\ybf|\dbf)}
\newcommand{\priorfp}{p(\ybf_T|\dbf_T)}

\newcommand{\appropto}{\mathrel{\vcenter{
  \offinterlineskip\halign{\hfil$##$\cr
    \propto\cr\noalign{\kern2pt}\sim\cr\noalign{\kern-2pt}}}}}

\begin{document}

\begin{frontmatter}
\title{Gradient-based Bayesian Experimental \\ Design for Implicit Models using \\ Mutual Information Lower Bounds}
\runtitle{Gradient-based BED for Implicit Models using MI Lower Bounds}

\begin{aug}
\author{\fnms{Steven} \snm{Kleinegesse}\thanksref{addr1}\ead[label=e1]{steven.kleinegesse@ed.ac.uk}}
\and
\author{\fnms{Michael U.} \snm{Gutmann}\thanksref{addr1}\ead[label=e3]{michael.gutmann@ed.ac.uk}}

\runauthor{S. Kleinegesse and M. Gutmann}

\address[addr1]{University of Edinburgh
    \printead{e1}, 
    \printead*{e3}
}


\end{aug}

\begin{abstract}
We introduce a framework for Bayesian experimental design (BED) with implicit models, where the data-generating distribution is intractable but sampling from it is still possible. In order to find optimal experimental designs for such models, our approach maximises mutual information lower bounds that are parametrised by neural networks. By training a neural network on sampled data, we simultaneously update network parameters and designs using stochastic gradient-ascent. The framework enables experimental design with a variety of prominent lower bounds and can be applied to a wide range of scientific tasks, such as parameter estimation, model discrimination and improving future predictions. Using a set of intractable toy models, we provide a comprehensive empirical comparison of prominent lower bounds applied to the aforementioned tasks. We further validate our framework on a challenging system of stochastic differential equations from epidemiology.
\end{abstract}



\end{frontmatter}

\section{Introduction} \label{sec:intro}

Parametric statistical models allow us to describe and study the behaviour of natural phenomena and processes. These models attempt to closely match the underlying process in order to simulate realistic data and, therefore, usually have a complex form and involve a variety of variables. Consequently, they tend to be characterised by intractable data-generating distributions (likelihood functions). Such models are referred to as $\emph{implicit models}$ and have an increasingly widespread use in the natural and medical sciences. Notable examples include high-energy physics~\citep{Agostinelli2003}, cosmology~\citep{Schafer2012}, epidemiology~\citep{Corander2017}, cell biology~\citep{Ross2017} and cognitive science~\citep{Palestro2018}.

Statistical inference is a key component of many scientific goals such as estimating model parameters, comparing plausible models and predicting future events. Unfortunately, because the likelihood function for implicit models is intractable, we have to revert to so-called likelihood-free, or simulation-based, inference to solve these downstream tasks~\citep[see][for a recent overview]{Cranmer2020}. Likelihood-free inference has gained much traction recently, with many methods leveraging advances in machine-learning~\citep[e.g.][]{Gutmann2016a, Lueckmann2017, Jarvenpaa2018, Chen2019, Papamakarios2019, Thomas2020, Jarvenpaa2021}. Ultimately, however, the quality of the statistical inference within a scientific downstream task depends on the data that are available in the first place. Because gathering data is often time-consuming and expensive, we should therefore aim to collect data that allow us to solve our specific task in the most efficient and appropriate manner possible.

Bayesian experimental design (BED) attempts to solve this problem by appropriately allocating resources in an experiment~\citep[see][for a review]{Ryan2016}. Scientific experiments generally involve controllable variables, called experimental designs $\dbf$, that affect the data gathering process. These might, for instance, be measurement times and locations, initial conditions of a dynamical system, or interventions and stimuli that are used to perturb a natural process. BED aims to find optimal experimental designs $\dbfopt$ that allow us to solve our scientific goals in the most efficient way. As part of this, we have to construct and optimise a utility function $U(\dbf)$ that indicates the value of an experimental design $\dbf$ according to the specific task at hand. In order to be fully-Bayesian, the utility function generally has to be a functional of the posterior distribution~\citep{Ryan2016}. This exacerbates the difficulty of BED for implicit models, as the involved posterior distributions are intractable. Proposing appropriate utility functions and devising methods to compute them efficiently for implicit models, with the help of likelihood-free inference, has been the focus of much present-day research in the field of BED~\citep[e.g.][]{Drovandi2013, Price2018, Overstall2020}.

A popular choice of utility function with deep roots in information theory is the $\emph{mutual information}$~\citep[MI;][]{Lindley1972} between two random variables, which describes the uncertainty reduction in one variable when observing the other. The MI is a standard utility function in BED for models with tractable likelihoods~\citep[e.g.][]{Ryan2003, Paninski2005, Overstall2017}, but has only recently started to gain traction in BED for implicit models~\citep[e.g.][]{Price2018, Kleinegesse2019, Foster2019, Kleinegesse2020a}. In the context of implicit models, the main difficulties that arise are 1) that estimating the posterior and obtaining samples from it via likelihood-free inference is expensive and 2) that gradients of the MI are generally not readily available, complicating its optimisation.

~\citet{Kleinegesse2020b} recently addressed the aforementioned problems by considering gradient-ascent on mutual information lower bounds. 
Importantly, they only considered one type of lower bound in their work, while there exist other bounds with different potentially helpful properties~\citep[see][for an overview]{Poole2019}. Moreover, their method only considers the specific task of parameter estimation, even though other research goals are often of equal practical importance.


In this paper, we extend the method of~\citet{Kleinegesse2020b} to other lower bounds and scientific aims, thereby devising a general framework for Bayesian experimental design based on mutual information lower bounds. Importantly, our framework does not require the use of variational distributions~\citep[as in][]{Foster2019}, but fully relies on parametrisation by neural networks. We believe that this increases practicality and generality, allowing scientists to use this method easily in a variety of settings. In short, our contributions are:
\begin{enumerate}
  \item We devise a BED framework that allows for the use of general lower bounds on mutual information in the context of implicit models, and 
  \item use it to perform BED for parameter estimation, model comparison and improving future predictions.
\end{enumerate}

\begin{figure}[!t]
    \centering
    \includegraphics[width=\linewidth]{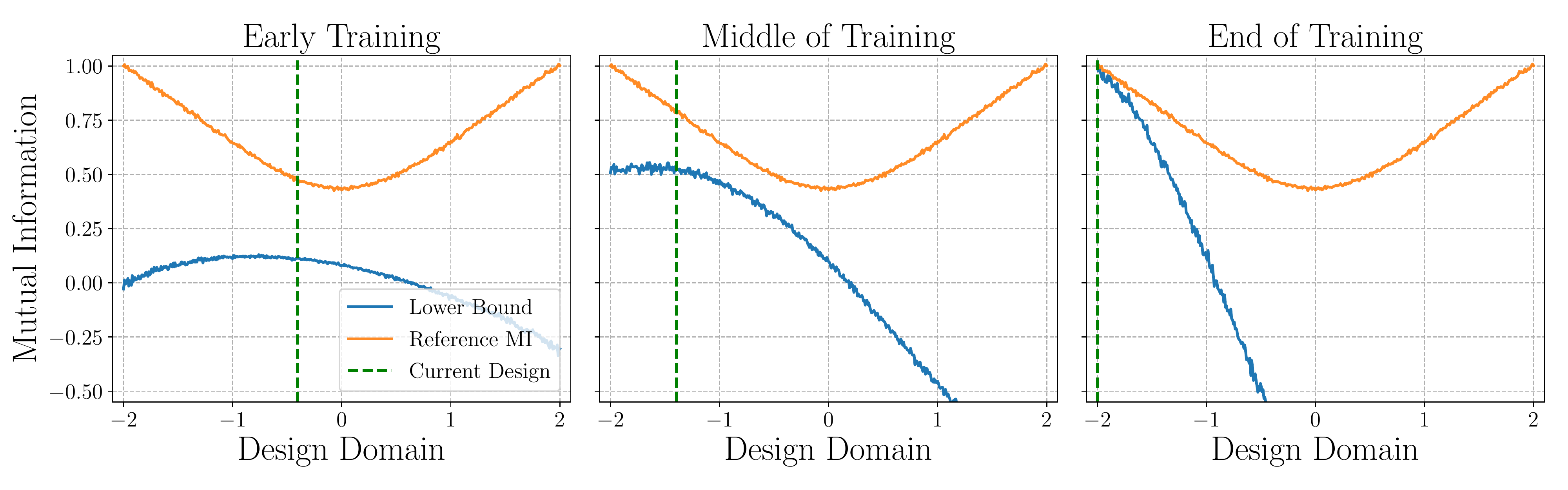}
    \vspace{-0.5cm}
    \caption[]{Visualisation of our proposed framework. During early times of training (left plot), the lower bound on mutual information is loose and the current design is far from the optimal design. We move closer to the optimal design while tightening the lower bound during the training process (middle plot). Finally, we find the optimal design and obtain a tight lower bound at the end of training (right plot). Note how we only spend resources on tightening the lower bound near the optimal design.}
    \label{fig:overview}
\end{figure}

Figure~\ref{fig:overview} provides a visualisation of our general framework, showcasing how training a neural network allows us to move towards optimal experimental designs. Our method optimises a parametrised lower bound with gradient-ascent, while simultaneously tightening it. This avoids spending resources on estimating the MI accurately for non-optimal designs, which can be seen in the right plot of Figure~\ref{fig:overview}, where the lower bound is only tight near the optimal design. Furthermore, once the optimal design has been found, our approach provides an amortised approximation of the posterior distribution, avoiding an additional costly likelihood-free inference step.
An animation corresponding to Figure~\ref{fig:overview} can be found in the public code repository that also includes the code to reproduce all results in this paper: \url{https://github.com/stevenkleinegesse/GradBED}.


We provide background knowledge of Bayesian experimental design with different scientific aims in Section~\ref{sec:background}. In Section~\ref{sec:methods} we present our BED framework for implicit models with different scientific aims and general lower bounds. This includes an introduction of parametrised mutual information lower bounds for BED, an explanation of how to optimise them and some practical guidance. In Section~\ref{sec:exp} we test our framework on a number of implicit models, followed by a discussion in Section~\ref{sec:concl}.

\clearpage

\section{Bayesian Experimental Design for different Aims} \label{sec:background}

In this section we provide background on Bayesian experimental design (BED) for different scientific aims using mutual information. Generally speaking, the aim of BED is to identify experimental designs $\dbf$ that allow us to achieve our scientific goals most rapidly. In order to do so, we first have to construct a utility function $U(\dbf)$ that tells us the worth of doing an experiment with design $\dbf \in \mathcal{D}$, where $\mathcal{D}$ specifies the domain of feasible experimental designs.
Optimal experimental designs $\dbfopt$ are then found by maximising the utility function $U(\dbf)$, i.e.
\begin{equation} \label{eq:opt}
  \dbfopt = \argmax_{\dbf \in \mathcal{D}} U(\dbf).
\end{equation}

Naturally, the choice of utility function $U(\dbf)$ is crucial, as it determines the optimal designs that are found. A popular utility function in BED is the mutual information (MI) between two random variables, which describes the uncertainty reduction, or expected information gain, in one variable when observing the other. It is a desirable quantitity because of its sensitivity to non-linearity and multi-modality, which other utilities cannot effectively deal with~\citep{Ryan2016, Kleinegesse2020a}. 

We here introduce the notion of a variable of interest $\vintbf$ that we wish to learn about by means of a scientific experiment. As such, our aim is to compute the MI $I(\vintbf;\ybf|\dbf)$ between the variable of interest $\vintbf$ and the observed data $\ybf$ at a particular design $\dbf$ (henceforth abbreviated by $\ybf|\dbf$). The MI utility function can then conveniently be adapted to different scientific aims by changing the variable of interest $\vintbf$ that is used in its computation~\citep{Ryan2016}. Its general form is given by
\begin{align}
U(\dbf) 
&= I(\vintbf; \ybf | \dbf) \label{eq:mi_general1} \\ 
&= \int \int \jointint \log \left( \frac{\jointint}{\margint \priorint} \right) \mathrm{d}\ybf \mathrm{d}\vintbf \label{eq:mi_general2}\\
&= \int \int \likeint \priorint \log \left( \frac{\postint}{\priorint} \right) \mathrm{d}\ybf \mathrm{d}\vintbf, \label{eq:mi_general3}
\end{align}
where we make the common assumption that the prior over the variable $\vintbf$ is unaffected by the designs $\dbf$, i.e. $p(\vintbf|\dbf) = \priorint$. In this form, the MI can also be reformulated as the expected Kullback-Leibler (KL) divergence~\citep{Kullback1951} between the posterior $\postint$ and prior $\priorint$. This interpretation means that we are looking for experimental designs resulting in data that maximally increase the difference between posterior and prior~\citep{Ryan2016}. In the case of implicit models, the data-generating distribution $\likeint$ and, hence, the posterior distribution $\postint$ are intractable, severely complicating the estimation and optimisation of Equation~\ref{eq:mi_general3}.


\paragraph{Parameter Estimation} The aim here is to find optimal designs that allow us to estimate the model parameters $\thetab$ most effectively. In this scenario, $\vintbf \rightarrow \thetab$ and the utility function is the MI $I(\thetab; \ybf | \dbf)$ between the model parameters $\thetab$ and data $\ybf|\dbf$,
\begin{equation}
U_{\text{PE}}(\dbf) = \int \int \like \prior \log \left( \frac{\post}{\prior} \right) \mathrm{d}\ybf \mathrm{d}\thetab.
\end{equation}
Parameter estimation is perhaps the most commonly-used scientific aim in BED for implicit models~\citep[e.g.][]{Drovandi2013, Hainy2016, Price2018, Kleinegesse2020a, Kleinegesse2020b}.

\paragraph{Model Discrimination} In this setting we wish to find designs that allow us to optimally discriminate between competing models. Each competing model is assigned a discrete model indicator $m$ that determines from which competing model the data is sampled, i.e. $\ybf \sim p(\ybf|m, \dbf)$. As such, we let $\vintbf \rightarrow m$ and define the utility function as the MI $I(m; \ybf | \dbf)$ between the model indicator $m$ and data $\ybf|\dbf$,
\begin{equation}
U_{\text{MD}}(\dbf) = \sum_m \int \likemd \priormd \log \left( \frac{\postmd}{\priormd} \right) \mathrm{d}\ybf.
\end{equation}
Generally, each competing model $m$ also has its own model parameters $\thetab_m$ that are needed for simulating data. In this particular case, where we are only concerned with discriminating between models, these model parameters are marginalised out. BED for model discrimination has recently received increased attention in the context of implicit models~\citep[e.g.][]{Dehideniya2018a, Hainy2019}.

\paragraph{Joint MD/PE} Combining the previous two tasks, we here wish to find designs that allow us to optimally discriminate between models $m$ and estimate a particular model's parameters $\thetab_m$. For this joint task we have $\vintbf \rightarrow (\thetab_m, m)$, with the utility function being the MI $I(\thetab_m, m; \ybf | \dbf)$ between $(\thetab_m, m)$ and simulated data $\ybf|\dbf$,
\begin{align}
U_{\text{MDPE}}(\dbf) = \sum_m \int \likemdpe \priormdpe \log \left( \frac{\postmdpe}{\priormdpe} \right) \mathrm{d}\ybf.
\end{align}
There have only been a few studies on joint parameter estimation and model discrimination in the context of BED for implicit models~\citep[e.g.][]{Dehideniya2018b}.

\paragraph{Improving Future Predictions} The premise of this setting is slightly different to the previous ones. We assume that we already know that we can perform an experiment with designs $\dbf_{T}$ at some time in the future, yielding (yet unknown) data $\ybf_{T}|\dbf_{T}$. However, we do have some budget to do a few initial experiments with designs $\dbf$ and observe initial data $\ybf$. Our aim is then to find the experimental designs that allow us to optimally predict the future observations $\ybf_{T}$ given the initial observations $\ybf$, i.e.~$\vintbf \rightarrow \ybf_{T}|\dbf_{T}$. The corresponding utility is the MI $I(\ybf_{T} | \dbf_T; \ybf | \dbf)$ between future observations $\ybf_{T}|\dbf_{T}$ and initial observations $\ybf$, conditioned on initial designs $\dbf$~\citep{Chaloner1995}:
\begin{align}
U_{\text{FP}}(\dbf) = \int \int \likefp \margfp \log \left( \frac{\likefp}{\priorfp} \right) \mathrm{d}\ybf \mathrm{d}\ybf_T,
\end{align}
where we have made the assumptions that $p(\ybf_T|\dbf, \dbf_T) = p(\ybf_T|\dbf_T)$ and $p(\ybf|\dbf, \dbf_T) = p(\ybf|\dbf)$. Importantly, the future design $\dbf_{T}$ stays constant during optimisation because we assume knowledge of this variable. Since we are only concerned with improving future predictions, we here marginalise out the model parameters $\thetab$. This utility function can also be interpreted as the expected KL divergence between the posterior predictive and prior predictive of future observations $\ybf_T$. As far as we are aware, there has only been one work on improving future predictions in the context of BED for implicit models~\citep[][]{Liepe2013}, but there has been some older work for explicit models where the likelihood is tractable~\citep[e.g.][]{Martini1984, Verdinelli1993}.

\begin{table}[t]
\begin{minipage}{0.9\textwidth}
\centering
\caption{Variable of interests and corresponding mutual information utilities for different scientific goals.}
\vspace{3mm} 
\begin{tabular}{lll} \toprule
Scientific Goal              & Variable of Interest $\vintbf$  & MI Utility          \\ \midrule
Parameter Estimation         & $\thetab$         & $I(\thetab; \ybf | \dbf)$.        \\
Model Discrimination         & $m$               & $I(m; \ybf | \dbf)$               \\
Joint MD/PE                  & $\thetab_m, m$    & $I(\thetab_m, m; \ybf | \dbf)$    \\
Improving Future Predictions & $\ybf_T | \dbf_T$ & $I(\ybf_T | \dbf_T; \ybf | \dbf)$ \\ \bottomrule
\end{tabular}
\label{tab:goals}
\end{minipage}
\end{table}

We summarise the above formulations in Table~\ref{tab:goals}; however, this list is by no means exhaustive. In fact, because we only have to change the variable of interest $\vintbf$ when computing the MI $I(\vintbf;\ybf|\dbf)$, we can easily incorporate a multitude of scientific goals. Furthermore, the formalism in Equation~\ref{eq:mi_general3} allows us to devise a general framework for BED with implicit models that is agnostic to a particular scientific goal.

\section{General BED Framework with Lower Bounds on MI} \label{sec:methods}

In this section we explain our general BED framework for implicit models based on mutual information lower bounds. We first provide motivation for using MI lower bounds and rephrase the BED problem accordingly. This is followed by examples of prominent lower bounds from literature. We identify common structures between them and explain how to use these to derive gradients with respect to designs, allowing us to solve the rephrased BED problem with gradient-based optimisation. Finally, we address a few technical difficulties and how to overcome them.

\subsection{Rephrasing the BED Problem}

Even though it is an effective and desirable metric of dependency, mutual information is notoriously difficult to estimate. This difficulty is exacerbated for implicit models, where we do not have access to the data-generating distribution in Equation~\ref{eq:mi_general3}. Recent work devised tractable mutual information estimators by combining variational bounds with machine-learning~\citep[see][for an overview]{Poole2019}. These methods generally work by using a neural network to parametrise a lower bound on mutual information, which is then tightened by training the neural network with gradient-ascent. There also exist some methods that use variational approximations to intractable distributions within lower bounds~\citep[e.g.][]{Donsker1983, Barber2003, Foster2019}. However, specifying a family of variational distributions carries the risk of mis-specification due to choosing an overly simple variational family. Parametrising non-linear functions such as neural networks is simpler than parametrising probability distributions and less prone to mis-specification.\footnote{Given that the neural network has enough capacity to capture the true distribution.} This difficulty is amplified in the context of implicit models, where we might have little knowledge about the underlying (intractable) distributions. Upper bounds on the mutual information generally require tractable distributions or variational approximations as well~\citep[e.g.][]{Poole2019, Cheng2020}.\footnote{We note that this might be because upper bounds on MI have been consistently understudied.} Thus, in this work we shall only be considering lower bounds on mutual information that are solely parametrised by neural networks.

Our overall goal is to maximise the MI $I(\vintbf; \ybf | \dbf)$ between a variable of interest $\vintbf$ and observed data $\ybf$ with respect to the design $\dbf$. Following the notation of~\citet{Belghazi2018}, we represent the scalar output of a neural network by $\NN$, where $\psib$ are the network parameters. We denote lower bounds on MI by $\mathcal{L}(\psib, \dbf)$; these depend on $\psib$ via the neural network $\NN$ and on $\dbf$ via the distributions over which expectations are taken. Our aim is then to maximise $\mathcal{L}(\psib, \dbf)$ with respect to $\psib$ and $\dbf$. The BED optimisation problem in Equation~\ref{eq:opt} becomes
\begin{equation} \label{eq:lbopt}
  \dbfopt = \argmax_{\dbf \in \mathcal{D}} \max_{\psib} \mathcal{L}(\psib, \dbf).
\end{equation}
As we maximise $\mathcal{L}(\psib, \dbf)$ with respect to $\psib$ we tighten the lower bound, while maximisation with respect to $\dbf$ allows us to find the optimal design. This optimisation problem is especially difficult for implicit models, as $\mathcal{L}(\psib, \dbf)$ generally involves expectations over distributions which depend on $\dbf$ in some complex manner. We will first discuss some prominent lower bounds from literature in Section~\ref{sec:lb} and then discuss how to optimise them in Section~\ref{sec:opt}.

\subsection{Prominent Lower Bounds} \label{sec:lb}

\paragraph{NWJ} The lower bound $\mathcal{L}_{\text{NWJ}}$ was first developed by Nguyen, Wainwright and Jordan~\citep[NWJ;][]{Nguyen2010} but is also known as $f$-GAN KL~\citep{Nowozin2016} and MINE-$f$~\citep{Belghazi2018}:
\begin{equation} \label{eq:nwj}
\mathcal{L}_{\text{NWJ}}(\psib, \dbf) \equiv  \Ej \left[ \NN \right] - e^{-1} \Em \left[ e^{\NN} \right].
\end{equation}
The optimal critic for this lower bound, i.e.~when it is tight, is $\NNopt = 1 + \log{\frac{\postint}{\priorint}}$. In other words, training a neural network $\NN$ with Equation~\ref{eq:nwj} as the objective function amounts to learning an amortised density ratio of the posterior to prior distribution. The NWJ bound has been shown to have a low bias but high variance~\citep{Poole2019, Song2020}.

\paragraph{InfoNCE} The InfoNCE lower bound $\mathcal{L}_{\text{NCE}}$ was introduced by~\citet{Oord2018} in the context of contrastive representation learning and takes the form
\begin{align}
\mathcal{L}_{\text{NCE}} (\psib, \dbf)
&\equiv  \mathbb{E} \left[ \frac{1}{K} \sum_{i=1}^K \log{\frac{e^{T_{\psib}(\vintbf_i, \ybf_i)}}{\frac{1}{K} \sum_{j=1}^K \left[ e^{T_{\psib}(\vintbf_j, \ybf_i)} \right] }} \right] \label{eq:info_1}, 
\end{align}
%
where the expectation is over $K$ independent samples from the joint distribution $\jointint$. The optimal critic for this lower bound is $\NNopt = \log{\likeint} + c(\ybf|\dbf)$ for the form shown above, where $c(\ybf|\dbf)$ is an indeterminate function. However, if we swap the indices of $\vintbf_j$ and $\ybf_i$ in the denominator, i.e.~sum out $\ybf$, the optimal critic becomes $\NNopt = \log{\postint} + c(\vintbf)$, where $c(\vintbf)$ is again indeterminate.
We use the version in Equation~\ref{eq:info_1}, as this allows us to obtain an unnormalised posterior that we can sample from, or normalise numerically in low dimensions (since $c(\ybf|\dbf)$ is fixed for a given real-world observation). The $\mathcal{L}_{\text{NCE}}$ lower bound has a low variance but, as noted by~\citet{Oord2018} and verified by~\citet{Poole2019}, it is upper-bounded by $\log{K}$, where $K$ is the batch-size in Equation~\ref{eq:info_1}, leading to a potentially large bias.

\paragraph{JSD} We are generally concerned with maximising the mutual information and not estimating it to a high accuracy. \citet{Hjelm2019} therefore argued to use the Jensen-Shannon divergence (JSD) as a proxy to mutual information.
They proposed to use a lower bound on the JSD between the joint distribution $\jointint$ and product of marginal distributions $\priorint \margint$, which takes the form
\begin{equation}
\mathcal{L}_{\text{JSD}}(\psib, \dbf) \equiv  \Ej \left[ -\text{sp}(-\NN) \right] - \Em \left[ \text{sp}(\NN) \right],
\end{equation}
where $\text{sp}(z) = \log(1 + e^z)$ is the softplus function
and the optimal critic of this lower bound is $\NNopt = \log \frac{\postint}{\priorint}$. The authors generally found the JSD lower bound to be more stable than other MI lower bounds. The JSD and KL divergence are both $f$-divergences and have a close relationship that can be derived analytically~\citep[see e.g.][]{Hjelm2019}. The authors also showed experimentally that distributions that lead to the highest JSD also lead to the highest MI, suggesting that maximising JSD is an appropriate proxy for maximising MI. If one is in fact concerned with accurate MI estimation as well, one could use the learned log density ratio in a MC estimation of the MI, or substitute it in any of the other MI lower bounds~\citep[as done in][]{Poole2019, Song2020}. We note that the JSD lower bound can be viewed as an expectation of the objective used in likelihood-free inference by ratio estimation~\citep[LFIRE;][]{Thomas2020}\footnote{See the supplementary material for more information on this relationship.}, which has been used before in the context of Bayesian experimental design~\citep{Kleinegesse2019, Kleinegesse2020a}, and is, like InfoNCE, related to noise-contrastive estimation~\citep[NCE;][]{Gutmann2012a} as discussed by \citet{Hjelm2019}.

We want to emphasise again that this work aims to showcase the use of general mutual information lower bounds, parametrised by neural networks, in Bayesian experimental design for different scientific aims. We do not claim that the aforementioned list of lower bounds is exhaustive and do not suggest that some bounds are particularly superior to others. We leave this investigation to more comprehensive studies focused exclusively on mutual information lower bounds~\citep[such as in][]{Poole2019}.

\clearpage

\subsection{Optimisation of Lower Bounds} \label{sec:opt}

In order to solve the rephrased BED problem in Equation~\ref{eq:lbopt}, we need to be able to optimise the lower bound $\mathcal{L}(\dbf, \psib)$ with respect to the neural network parameters $\psib$ and the experimental designs $\dbf$, regardless of the choice of lower bound. In the interest of scalability~\citep{Spall2003}, we here optimise both of these with gradient-based methods.

Looking at MI lower bounds in literature, including the prominent ones shown in Section~\ref{sec:lb}, we can identify a few similar structures that allow us to generalise gradient-based optimisation. First, lower bounds generally involve expectations over the joint distribution and/or the product of marginal distributions. Second, these expectations tend to contain non-linear, differentiable functions of the neural network output. In other words, lower bounds usually involve expectations $\Ej \left[ f(\NN) \right]$ and/or $\Em \left[ g(\NN) \right]$, where $f:\mathbb{R}\rightarrow \mathbb{R}$ and $g:\mathbb{R}\rightarrow\mathbb{R}$ are non-linear, differentiable functions. For instance, for the NWJ lower bound in Equation~\ref{eq:nwj} we have $f(z)=z$ and $g(z)=e^{z-1}$. By deriving gradients of these expectations with respect to $\psib$ and $\dbf$ we can then easily derive gradients for all prominent lower bounds listed in the Section~\ref{sec:lb}, as well as
other ones that share the same structure. 

Fortunately, gradients of the aforementioned expectations with respect to the network parameters $\psib$ are straightforward, as the involved distributions do not depend on them. This means that we can simply pull the gradient operator $\nabla_{\psib}$ inside the expectations and apply the chain rule, yielding
\begin{align}
\nabla_{\psib} \Ej \left[ f(\NN) \right] &= \Ej \left[ f^{\prime}(T)\Bigr|_{T=\NN} \nabla_{\psib} \NN \right] \quad \text{and} \\
\nabla_{\psib} \Em \left[ g(\NN) \right] &= \Em \left[ g^{\prime}(T)\Bigr|_{T=\NN} \nabla_{\psib} \NN \right],
\end{align}
where $f^{\prime}(T) = \partial f(T) / \partial T$ and $g^{\prime}(T) = \partial g(T) / \partial T$. The first factor inside the expectations only depends on the form of the non-linear functions, while the second factor, the gradients of the network output with respect to its parameters, is generally computed with automatic differentiation, i.e.~back-propagation. The expectations can then be approximated as sample averages with $N$ samples from the corresponding distributions.

Gradients with respect to the designs $\dbf$ are more complicated, as the intractable distributions over which expectations are taken depend on $\dbf$. For instance, we cannot compute $\nabla_{\dbf} \Ej \left[ f(\NN) \right]$ in the same way that we computed the previous gradients with respect to $\psib$, because the joint distribution $\jointint$ (and therefore its gradient) is intractable. Similarly, we cannot use score-function estimators~\citep{Kleijnen1996} to approximate these gradients, as they require an analytic derivative of the log densities.


However, the pathwise gradient estimator lends itself to our setting with implicit models~\citep[see][for a review]{Mohamed2020}. Indeed, like implicit models, this estimator assumes that sampling from the data-generation distribution $\likeint$ is exactly the same as sampling from a base distribution $p(\bm{\epsilon})$ and then using that noise sample as an input to a non-linear, deterministic function $\mathbf{h}(\bm{\epsilon}; \vintbf, \dbf)$~\citep{Mohamed2020}, called the sampling path, i.e.\footnote{For instance, sampling a Gaussian random variable from $\mathcal{N}(y; \mu, \sigma^2)$ is exactly the same as first sampling noise from a standard normal $\mathcal{N}(\epsilon; 0, 1)$ and then computing $y = \mu + \epsilon \sigma$.}
\begin{equation} \label{eq:samplepath}
\ybf \sim \likeint \iff \ybf = \mathbf{h}(\bm{\epsilon}; \vintbf, \dbf), \quad \bm{\epsilon} \sim p(\bm{\epsilon}).
\end{equation}
%
This then allows us to invoke the law of the unconscious statistician~\citep[LOTUS; e.g.][]{Grimmett2001} and, for instance, rephrase the expectation over the joint $\jointint$ in terms of the base distribution $p(\bm{\epsilon})$ and the prior over $p(\vintbf)$,
\begin{equation} \label{eq:exp_1}
  \Ej \left[ f(\NN) \right] = 
  \mathbb{E}_{p(\bm{\vint})p(\bm{\epsilon})} \left[ f(T_{\bm{\psi}}(\bm{\vint}, \mathbf{h}(\bm{\epsilon}; \vintbf, \dbf))) \right],
\end{equation}
where we have factorised the joint distribution as $\jointint = \likeint \priorint$, assuming that $\priorint$ is unaffected by $\dbf$. 

For the expectation over the product of marginals we need to perform an additional step. Equation~\ref{eq:samplepath} specifies the sampling path $\mathbf{h}(\bm{\epsilon}; \vintbf, \dbf)$ of the data-generation distribution, but we require the sampling path of the marginal data $\ybf \sim p(\ybf | \dbf)$. This means that we need to express the marginal as an expectation of the data-generating distribution over the prior, i.e. $p(\ybf|\dbf) = \mathbb{E}_{p(\widetilde{\vintbf})} [p(\ybf|\widetilde{\vintbf}, \dbf)]$, where $p(\widetilde{\vintbf})$ is exactly the same distribution as $\priorint$. We can then specify $\ybf = \mathbf{h}(\bm{\epsilon}; \widetilde{\vintbf}, \dbf)$ and invoke LOTUS again, yielding the second rephrased expectation:
\begin{equation} \label{eq:exp_2}
  \Em \left[ g(\NN) \right] = 
  \mathbb{E}_{p(\vintbf)p(\widetilde{\vintbf})p(\bm{\epsilon})} \left[ g(T_{\bm{\psi}}(\bm{\vint}, \mathbf{h}(\bm{\epsilon}; \widetilde{\vintbf}, \dbf))) \right].
\end{equation}

The rephrased expectations in Equations~\ref{eq:exp_1}~--~\ref{eq:exp_2} are now taken over distributions that do not directly depend on the designs $\dbf$, allowing us to easily take gradients with respect to $\dbf$. In the interest of space, let us define $T = T_{\bm{\psi}}(\bm{\vint}, \mathbf{h}(\bm{\epsilon}; \vintbf, \dbf))$ and $\widetilde{T} = T_{\bm{\psi}}(\bm{\vint}, \mathbf{h}(\bm{\epsilon}; \widetilde{\vintbf}, \dbf))$. The required gradients are then
\begin{align}
\nabla_{\dbf} \Ej \left[ f(T) \right] &= 
\mathbb{E}_{p(\bm{\vint})p(\bm{\epsilon})} \left[ f^{\prime}(T) \nabla_{\dbf} T \right] \quad \text{and} \label{eq:grad_final_1} \\
\nabla_{\dbf} \Em \left[ g(T) \right] &= 
\mathbb{E}_{p(\bm{\vint})p(\widetilde{\vintbf})p(\bm{\epsilon})} \left[ g^{\prime}(\widetilde{T}) \nabla_{\dbf} \widetilde{T} \right], \label{eq:grad_final_2}
\end{align}
where the derivatives $f^{\prime}(T) = \partial f(T) / \partial T$ and $g^{\prime}(\widetilde{T}) = \partial g(\widetilde{T}) / \partial \widetilde{T}$ depend on the lower bound in question.
The $\nabla_{\dbf} T$ and $\nabla_{\dbf} \widetilde{T}$ factors in the above equations are the gradients of the network output with respect to designs, which is the most difficult technical part of our method. We discuss different approaches and caveats to computing these gradients in detail in Section~\ref{sec:samp_path}.

We next briefly consider a special setting where the implicit model is defined such that we can separate the data generation into a `known' observational process\footnote{With `known', we mean that the model of the observational process is known analytically in closed form.} $p(\ybf|\vintbf, \dbf, \bm{z})$ and a differentiable, stochastic latent process $p(\bm{z}|\dbf, \vintbf)$, where $\bm{z}$ is a latent variable. In this case we can apply a score-function estimator to the observational process, because we assume knowledge of the likelihood $p(\ybf|\vintbf, \dbf, \bm{z})$, and a pathwise gradient estimator to the latent process. In a similar manner to Equation~\ref{eq:exp_1}, we can, for instance, rephrase the expectation of $f(\NN)$ over the joint distribution as
\begin{align} \label{eq:exp_latent}
  \Ej \left[ f(\NN) \right] 
  &= \mathbb{E}_{p(\ybf|\vintbf, \dbf, \bm{z}) p(\bm{z}|\dbf, \vintbf) p(\bm{\vint})} \left[ f(\NN) \right] \\
  &= \mathbb{E}_{p(\ybf|\vintbf, \dbf, \mathbf{h}(\bm{\epsilon}; \vintbf, \dbf)) p(\bm{\epsilon}) p(\bm{\vint}) } \left[ f(\NN) \right],
\end{align}
where in this case $\bm{\epsilon}$ is the noise random variable that defines the sampling path of the latent stochastic variable $\bm{z} = \mathbf{h}(\bm{\epsilon}; \vintbf, \dbf)$. This allows us to pull the gradient operatior $\nabla_{\dbf}$ into the required expectations; see the supplementary material for more details on the resulting estimator. This rephrasing is useful because 1) we can inject extra knowledge into the method as we know parts of the data generation process and 2) this also allows us to deal with some implicit models that generate discrete data (in which case the full gradient with respect to $\dbf$ is undefined). For instance, in experiments later in the paper, we present an implicit model where the latent process is the solution of a stochastic differential equation, which is differentiable by nature, and the observational process is a known discrete Poisson process.

Lastly, we apply the aforementioned methodology of computing gradients to all of the prominent lower bounds listed in Section~\ref{sec:lb}, as they only effectively differ in what kind of non-linear functions $f$ and $g$ they use, yielding the gradients shown in Table~\ref{tab:grads}. A similar table for the special setting of knowing the observational process is shown in the supplementary materials. Furthermore, we provide an overview of the discussed framework of Bayesian experimental design for implicit models using lower bounds on mutual information in Algorithm~\ref{algo:mibed}.
\begin{table}[!t]
\begin{minipage}{0.9\textwidth}
\centering
\caption{Gradients of several lower bounds with respect to experimental designs. We define $T = T_{\bm{\psi}}(\bm{\vint}, \mathbf{h}(\bm{\epsilon}; \vintbf, \dbf))$, $\widetilde{T} = T_{\bm{\psi}}(\bm{\vint}, \mathbf{h}(\bm{\epsilon}; \widetilde{\vintbf}, \dbf))$ and $T_{ij} = T_{\bm{\psi}}(\bm{\vint}_j, \mathbf{h}(\bm{\epsilon}_i; \vintbf_i, \dbf))$. We also use the shorthand $P=p(\vintbf)p(\bm{\epsilon})$ and $Q=p(\vintbf)p(\widetilde{\vintbf})p(\bm{\epsilon})$. Here, $\sigma(T)$ is the logistic sigmoid function. See the supplementary material for detailed derivations and a corresponding table for the case where the observation model is analytically tractable.}
\vspace{3mm}
\begin{tabular}{lll} \toprule
Lower Bound & Gradients with respect to designs \\ \midrule
NWJ         & $\mathbb{E}_{P} \left[ \nabla_{\dbf} T \right]  - e^{-1} \mathbb{E}_{Q} \left[ e^{\widetilde{T}} \nabla_{\dbf} \widetilde{T} \right]$ \\
InfoNCE     & $\mathbb{E}_{P^K} \left[ \frac{1}{K} \sum_{i=1}^K \frac{\sum_{j=1}^K e^{T_{ij}} (\nabla_{\dbf} T_{ii} - \nabla_{\dbf} T_{ij})}{\sum_{j=1}^K e^{T_{ij}}}\right]$ \\ 
JSD         & $\mathbb{E}_{P} \left[ \sigma(-T) \nabla_{\dbf} T \right] - \mathbb{E}_{Q} \left[ \sigma(\widetilde{T}) \nabla_{\dbf} \widetilde{T} \right]$ \\ \bottomrule
\end{tabular}
\label{tab:grads}
\end{minipage}
\end{table}
\begin{algorithm}[!t]
\caption{Gradient-Based BED for Implicit Models using MI Lower Bounds}\label{algo:mibed}
\begin{algorithmic}[1]
\Input Implicit model sampling path $\mathbf{h}(\bm{\epsilon}; \vintbf, \dbf)$, neural network $\NN$, MI lower bound $\mathcal{L}(\psib, \vintbf)$, optimisers for both $\psib$ and $\dbf$, number of samples $N$
\Output Optimal design $\dbfopt$, trained neural network $T_{\psib^{\ast}}(\vintbf, \ybf)$, MI estimate $\mathcal{L}(\psib^{\ast}, \dbfopt)$
\item[] \vspace{-0.15cm}
\State Sample from the prior over the variable of interest: $\vintbf^{(i)} \sim \priorint$ for $i=1, \dots, N$
\State Randomly initialise the neural network parameters $\psib$
\State Randomly initialise the experimental designs $\dbf$
\While {$\mathcal{L}(\psib, \dbf)$ not converged}
  \State {Sample noise from the base distribution, i.e.~$\bm{\epsilon}^{(i)} \sim p(\bm{\epsilon})$ for $i=1, \dots, N$}
  \State {Obtain joint data via the sampling path $\ybf^{(i)} = \mathbf{h}(\bm{\epsilon}^{(i)}, \vintbf^{(i)}, \dbf)$ for $i=1, \dots, N$}
  \State {Randomly shuffle data to obtain marginal samples $\{\widetilde{\ybf}^{(i)}\}_{i=1}^N$}
  \State {Compute network output with joint samples: $T_{\psib}(\vintbf^{(i)}, \ybf^{(i)})$ for $i=1, \dots, N$}
  \State {Compute network output with marginal samples: $T_{\psib}(\vintbf^{(i)}, \widetilde{\ybf}^{(i)})$ for $i=1, \dots, N$}
  \State {Compute a sample average of the MI lower bound $\mathcal{L}(\psib, \dbf)$}
  \State {Estimate gradients with respect to $\psib$ and $\dbf$}
  \State {Update $\psib$ and $\dbf$ using two separate optimisers}
\EndWhile
\end{algorithmic}
\end{algorithm}

\subsection{Tackling the Sampling Path Gradients} \label{sec:samp_path}

Recall the gradients of the expectations in Equations~\ref{eq:grad_final_1}~--~\ref{eq:grad_final_2},which involve the gradient of the network output with respect to designs, i.e.$\nabla_{\dbf} T_{\bm{\psi}}(\bm{\vint}, \mathbf{h}(\bm{\epsilon}; \vintbf, \dbf))$. Below we discuss several approaches to computing, or approximating, this gradient.

\pagebreak

\paragraph{Automatic Differentiation} As mentioned previously, the gradients of lower bounds with respect to $\psib$ are already computed with automatic differentiation in a standard neural network training fashion. We can similarly use automatic differentiation to compute gradients with respect to $\dbf$ if the implicit model is written in a differential programming framework, such as JAX~\citep{JAX2018}, PyTorch~\citep{PyTorch2019} or TensorFlow~\citep{Tensorflow2015}. This allows us to attach the simulator model to the computation graph of the framework and back-propagate from evaluations of the lower bound to the experimental designs $\dbf$. This means that we do not have to explicitly provide gradients, which is helpful if the implicit model is extremely complex, and allows us to fully-utilise parallelisation with GPUs. Furthermore, we expect the usefulness of this aspect, and the capability of our method, to grow with the development of automatic differentiation frameworks, which are already essential in machine learning.

\paragraph{Manual Gradient Computation} In some situations we may need to manually compute the required gradients, for example when the implicit model is not written in a differential programming framework. To do so, we first apply the chain rule to $\nabla_{\dbf} T_{\bm{\psi}}(\bm{\vint}, \mathbf{h}(\bm{\epsilon}; \vintbf, \dbf))$, yielding
\begin{equation}
\nabla_{\dbf} T_{\bm{\psi}}(\bm{\vint}, \mathbf{h}(\bm{\epsilon}; \vintbf, \dbf)) = 
\mathbf{J_y}^{\top} \nabla_{\ybf} \NN \bigr|_{\ybf = \mathbf{h}(\bm{\epsilon}; \vintbf, \dbf)}. \label{eq:gradT_1}
\end{equation} 
The factor $\nabla_{\ybf} \NN$ is the derivative of the neural network output with respect to its inputs, which can be readily obtained via automatic differentiation in most machine-learning frameworks. $\mathbf{J_y}$ is the Jacobian matrix, defined by ~$(\mathbf{J_{y}})_{ij} = \partial y_i / \partial d_j$, and contains the gradients of the sampling path with respect to the designs. We either need to assume knowledge of the sampling path gradients or find ways to approximate them.
This includes common settings where the data generation can be written down in one line, e.g.~a linear model with a combination of noises from different non-Gaussian sources, or we have access to the ordinary, or stochastic, differential equations that govern data-generation.


\paragraph{Gradient-Free Optimisation} While we can facilitate gradient-based optimisation for a variety of implicit models, there might be situations in which we can neither exactly compute nor approximate the required Jacobian-vector product through any of the aforementioned ways. For instance, sampling path gradients might be undefined if the data generation process involves any discrete variables that we need to differentiate and we cannot separate it into a known observational process and a latent process.
In such situations we can only resort to gradient-free optimisation methods for the updates of $\dbf$. 
Examples of gradient-free optimisation techniques for BED with implicit models are grid search, random search, evolutionary algorithms~\citep[as done in e.g.][]{Zhang2021} and Bayesian optimisation~\citep[as done in e.g.][]{Kleinegesse2020b}. We note that in some situations we might be able to use continuous relaxations of discrete variables instead, which we shall leave for future work.
While the gradient-free methods allow us to deal with a larger class of implicit models, they do not scale well with the dimensionality of experimental designs $\dbf$~\citep{Spall2003}. 
As such, we do not consider the gradient-free setting in this work, but refer the reader to~\citet{Kleinegesse2020b} for more detailed explanations and experiments for such a setting.


\section{Experiments} \label{sec:exp}

In this section we demonstrate our approach to BED for implicit models using MI lower bounds. First, we consider a toy example that consists of a set of models with linear, logarithmic and square-root responses. In a comprehensive comparison, we apply all prominent lower bounds presented in Section~\ref{sec:lb} (NWJ, InfoNCE and JSD) to all scientific goals presented in Section~\ref{sec:background} (PE, MD, MD/PE and FP). Second, we consider a setting from epidemiology where the latent process is the solution to a stochastic differential equation and the discrete observational process is analytically tractable. We here only apply the JSD lower bound on the PE and MD tasks. 
Code to reproduce results can be found here: \url{https://github.com/stevenkleinegesse/GradBED}.


\subsection{Toy Models}

We here assume that an observable variable $y \in \mathbb{R}$ depends on an experimental design $d\in [-2, 2]$ in either a linear ($m=1$), logarithmic ($m=2$) or square-root ($m=3$) manner, where the variable $m \in \{1, 2, 3\}$ is the model indicator. Each model $m$ is also governed by two model parameters $\thetab_m = (\theta_{m,0}, \theta_{m,1})$, e.g.~the offset and slope in the case of the linear model. We assume a limited budget of $10$ measurements, which means that we have to construct a 10-dimensional design vector $\dbf = (d_1, \dots, d_{10})$ with a corresponding data vector $\ybf = (y_1, \dots, y_{10})$. We include two separate sources of noise, Gaussian noise $\mathcal{N}(\epsilon; 0,1)$ and Gamma noise $\Gamma(\nu; 2, 2)$, which leads to all toy models having likelihood functions that do not admit a closed-form expression.
The sampling paths are given by
\begin{equation} \label{eq:toy_models}
\ybf = 
\begin{cases}
\theta_{1,0} \mathbf{1} + \theta_{1,1} \dbf + \bm{\epsilon} + \bm{\nu} & \text{if } m = 1, \\
\theta_{2,0} \mathbf{1} + \theta_{2,1} \log (|\dbf|) + \bm{\epsilon} + \bm{\nu} & \text{if } m = 2, \\
\theta_{3,0} \mathbf{1} + \theta_{3,1} \sqrt{|\dbf|} + \bm{\epsilon} + \bm{\nu} & \text{if } m = 3,
\end{cases}
\end{equation}
where $\bm{\epsilon} = (\epsilon_1, \dots, \epsilon_{10})$ and $\bm{\nu} = (\nu_1, \dots, \nu_{10})$ are iid samples from a Gaussian and Gamma noise source, respectively, $\mathbf{1}$ denotes a 10-dimensional vector of ones and $|\cdot|$ implies taking the element-wise absolute value. To ensure numerical stability in the logarithmic model, we clipped $|\dbf|$ by $10^{-4}$ from below, i.e. $|\dbf| \equiv \max(|\dbf|, 10^{-4})$. We generally use a Gaussian prior $\mathcal{N}(\thetab_m; \mathbf{0}, 3^2 \mathds{1})$ over the model parameters and a discrete uniform prior $\mathcal{U}(m)$ over the model indicators.

Comprehensive information about neural network architectures, hyper-parameters and implementation details can be found in the supplementary material. Throughout this section we compare final MI estimates and posterior distributions to reference values and distributions. Information about how these are computed for all scientific tasks can also be found in the supplementary materials.

\subsubsection{Parameter Estimation for the Linear Model}

\begin{figure}[!t]
    \centering
    \includegraphics[width=\linewidth]{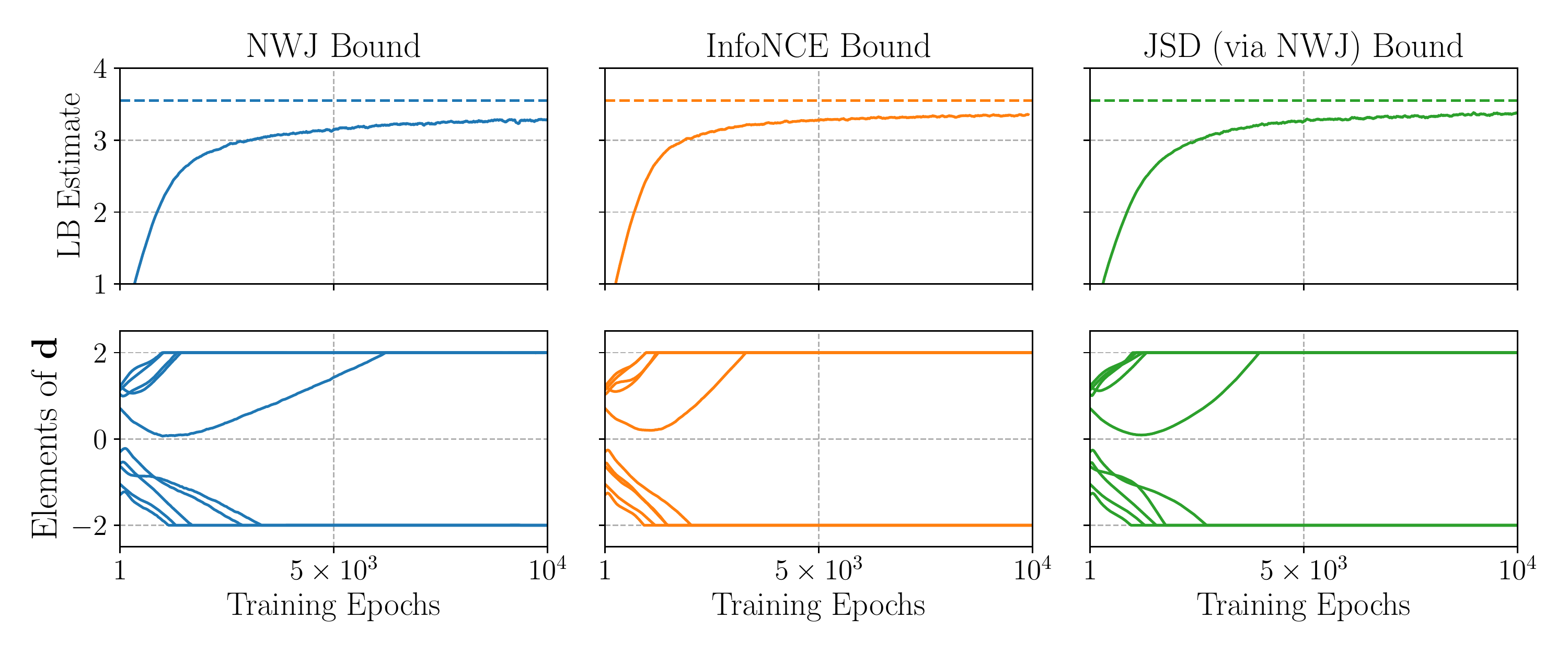}
    \vspace{-0.5cm}
    \caption[]{PE results for the linear toy model. Shown are training curves for the NWJ, InfoNCE and JSD (via NWJ) lower bounds. The top row shows lower bound estimates as a function of training epochs, with the dotted line being the reference MI value, and the bottom row shows the elements of the design vector as it is being updated.}
    \label{fig:toy_pe_bounds}
\end{figure}

We first consider the scientific task of parameter estimation (PE) for the linear toy model ($m=1$ in Equation~\ref{eq:toy_models}), where the aim is to estimate the model parameters $\thetab_1 = (\theta_{1,0}, \theta_{1,1})$, i.e.~the slope and offset of a straight line, respectively.
Figure~\ref{fig:toy_pe_bounds} shows the training results for the PE task, using the NWJ (left column), InfoNCE (middle column) and JSD (right column) lower bounds.
As explained in Section~\ref{sec:lb}, we use the density ratio learned by maximising the JSD lower bound as an input to the NWJ lower bound, in order to get an actual lower bound on the mutual information, which can be seen in the right column.
All lower bounds converge smoothly to a final value that is close to the reference MI value, with all lower bounds finding the same optimal design $\dbf^\ast$, whose elements are equally clustered at $d=-2$ and $d=2$. Experimental designs at the boundaries of the design domain are useful because that is where the signal-to-noise ratio is highest. Intuitively, this means that our data contains more signal than noise, allowing us to better measure the effect of model parameters on data.
Interestingly, there is no noticeable difference in the convergence behaviour of each lower bound (they all use the same neural network and design initialisations). 

%
\begin{table}[!t]
\begin{minipage}{0.9\textwidth}
\centering
\caption{Toy model average MI estimates ($\pm$ standard error) for all scientific tasks, using the NWJ, InfoNCE and JSD (via NWJ) lower bounds on MI (higher values are better). Also shown are average reference MI estimates.
}
\vspace{3mm} 
\begin{tabular}{lcccc} \toprule
MI LB & PE                   & MD                    & MD/PE                & FP                   \\ \midrule
NWJ   & 3.43 {$\pm$} 0.06    & 0.726 {$\pm$} 0.007     & 3.72 {$\pm$} 0.06    & 1.33 {$\pm$} 0.01    \\
InfoNCE   & 3.36 {$\pm$} 0.06    & 0.725 {$\pm$} 0.027     & 3.80 {$\pm$} 0.07    & 1.33 {$\pm$} 0.04    \\
JSD   & \B 3.48 {$\pm$} 0.05 & \B 0.730 {$\pm$} 0.007  & \B 3.95 {$\pm$} 0.08 & \B 1.34 {$\pm$} 0.01 \\ \midrule
Ref.  & 3.55 {$\pm$} 0.04    & 0.737 {$\pm$} 0.007     & 3.97 {$\pm$} 0.02    & 1.34 {$\pm$} 0.02    \\ \bottomrule
\end{tabular}
\label{tab:toy_mi}
\end{minipage}
\end{table}

The PE column in Table~\ref{tab:toy_mi} shows the final MI estimates for each lower bound, evaluated on several synthetic validation sets, with the JSD lower bound having the highest estimate. This is intriguing because, while the gradients are computed with the JSD lower bound, we evaluate it with NWJ lower bound and, yet, its estimate is higher than that for the actual NWJ lower bound. 

\begin{figure}[!t]
    \centering
    \includegraphics[width=\linewidth]{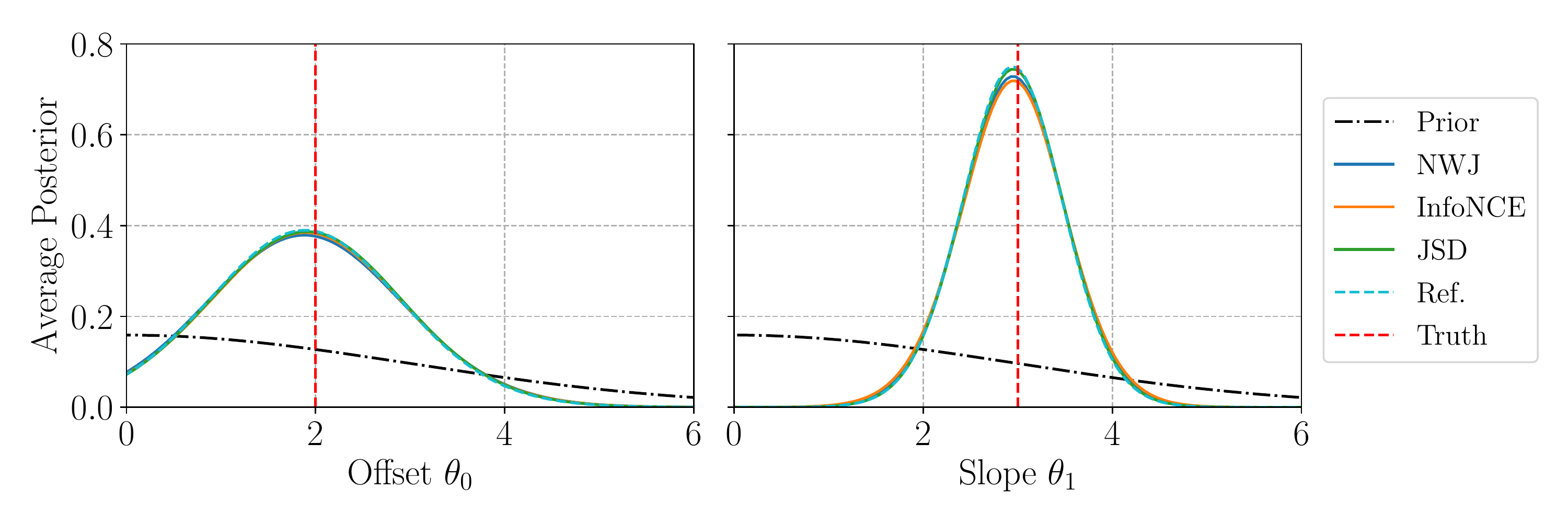}
    \vspace{-0.7cm}
    \caption[]{PE posterior results for the linear toy model. Shown are average marginal posteriors for all lower bounds, including the prior, the ground truth and the average reference posterior. Note how all lower bounds produce similar posteriors.}
    \label{fig:toy_pe_post}
\end{figure}

As explained in Section~\ref{sec:lb}, we can use the trained neural network to directly compute a posterior distribution. We first generate synthetic `real-world' data $\ybf^\ast$ that was sampled with some ground-truth $\thetab_{1, \text{true}} = (2, 3)$ at the optimal design $\dbf^\ast$, i.e.~$\ybf^\ast \sim p(\ybf|\dbf^\ast, \thetab_{1,\text{true}})$. The prior distribution, the trained neural network and the observation can then be used to compute the posterior $p(\thetab|\ybf^\ast, \dbf^\ast)$ according to Section~\ref{sec:lb}. The top row in Figure~\ref{fig:toy_pe_post} shows the marginal posteriors of the offset $\theta_{1,0}$ and slope $\theta_{1,1}$ for each lower bound, averaged over $5{,}000$ `real-world' observations. Also shown are average reference posteriors computed with the same observations.\footnote{See the supplementary material for how we compute these.} All posterior distributions cover the ground truth well, and their MAP estimate is close to the ground truth.
However, there is no noticeable difference between any of the shown posterior distributions. The relative performance of each lower bound is more apparent in Table~\ref{tab:toy_kl}, where we show the average KL-Divergence between estimated and reference posteriors. The JSD lower bound results in posterior distributions that are closest to the reference posteriors.
\begin{table}[!t]
\begin{minipage}{0.9\textwidth}
\centering
\caption{Toy model average KL-Divergences ($\pm$ standard error) between estimated and reference posteriors for all scientific tasks and all lower bounds (lower values are better). 
}
\vspace{3mm} 
\begin{tabular}{lcccc} \toprule
MI LB     & PE                     & MD                     & MD/PE                  & FP \\ \midrule
NWJ       & 0.079 {$\pm$} 0.003    & 0.016 {$\pm$} 0.001    & 0.728 {$\pm$} 0.010    & 0.009 {$\pm$} 0.001      \\
InfoNCE   & 0.107 {$\pm$} 0.003    & 0.018 {$\pm$} 0.001    & 0.053 {$\pm$} 0.001    & 0.012 {$\pm$} 0.001      \\
JSD       & \B 0.028 {$\pm$} 0.001 & \B 0.009 {$\pm$} 0.001 & \B 0.026 {$\pm$} 0.001 & \B 0.005 {$\pm$} 0.001   \\
\bottomrule
\end{tabular}
\label{tab:toy_kl}
\end{minipage}
\end{table}

\subsubsection{Model Discrimination}

Next, we consider the task of model discrimination (MD), where the aim is to distinguish between competing models and the variable of interest is the model indicator $m \in \{1, 2, 3\}$. 
To sample data $\ybf \sim p(\ybf | \dbf, m)$, as required in Algorithm~\ref{algo:mibed}, we need to marginalise over the model parameters $\thetab_m$. To do so, we first obtain prior samples, use these to sample data $\ybf$ from the sampling path in Equation~\ref{eq:toy_models} and then simply discard the $\thetab_m$ samples. 

\begin{figure}[!t]
    \centering
    \includegraphics[width=\linewidth]{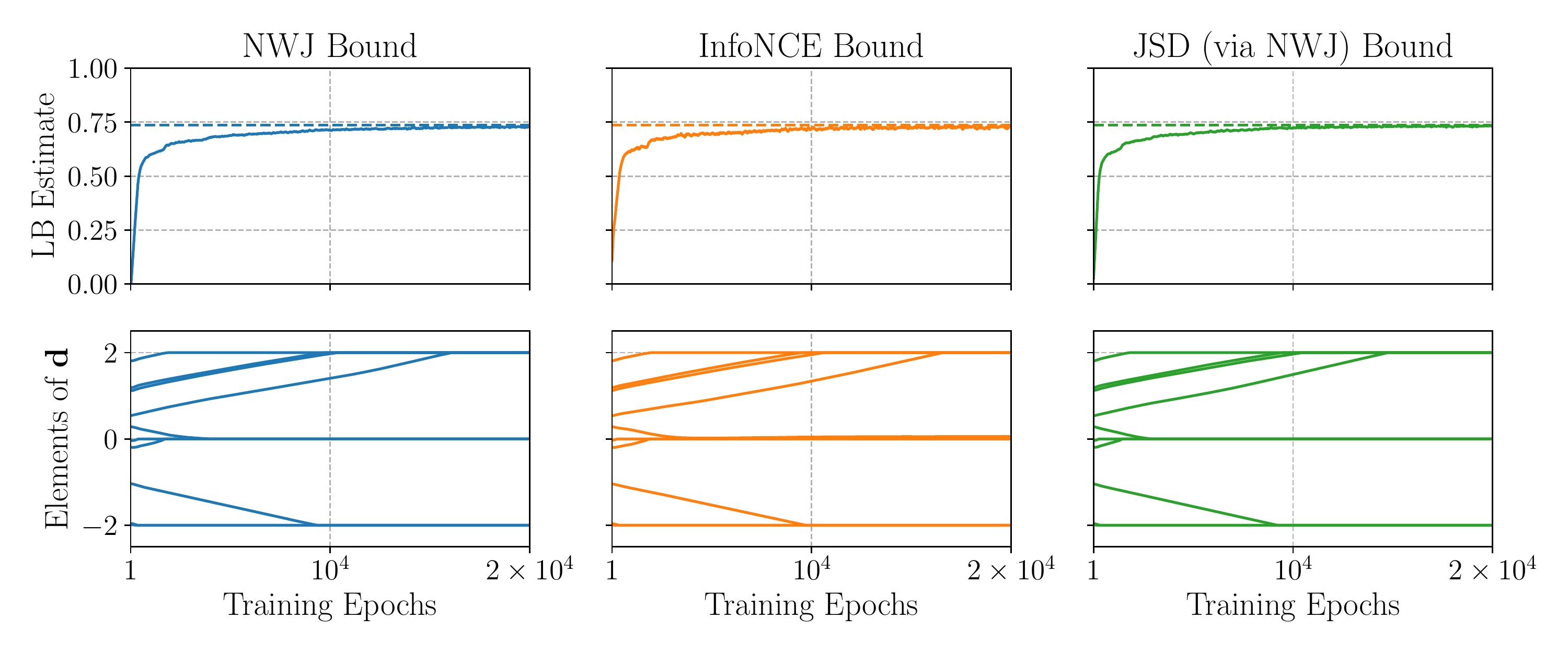}
    \vspace{-0.5cm}
    \caption[]{MD results for the set of toy models. Shown are training curves for the NWJ, InfoNCE and JSD (via NWJ) lower bounds. The top row shows lower bound estimates as a function of training epochs, with the dotted line being the reference MI value, and the bottom row shows the elements of the design vector as it is being updated.}
    \label{fig:toy_md_bounds}
\end{figure}

Figure~\ref{fig:toy_md_bounds} shows the training results for the MD task, using the NWJ (left column), InfoNCE (middle column) and JSD (right column) lower bounds. Similar to the PE task, we evaluate the JSD bound by using the learned density ratio as an input to the NWJ lower bound. All lower bounds have a similar convergence behaviour and lead to final MI estimates that are close to the reference MI value. Similar to the PE tasks, all lower bounds find the same optimal design $\dbf^\ast$, which consists of $3$ elements at $d=-2$, $4$ elements at $d=0$ and $4$ elements at $d=2$. The additional design cluster around $d=0$, as compared to the PE task, is useful for model discrimination because of the large response of the logarithmic model ($m=2$). Although we clipped $|\dbf|$ from below by $10^{-4}$ when $m=2$, this still means that the response of the logarithmic model is significantly larger than that of the linear model ($m=1$) and the square-root model ($m=3$) as $d \rightarrow 0$. This allows us to determine with relative ease whether or not observed data was generated from the logarithmic model. Similarly, the other two clusters at the boundaries of the design domain are helpful because that is where the data distributions of all models are most different, which is helpful in distinguishing between them.\footnote{See the supplementary material for a figure showing this.}

Evaluations of the lower bounds on validation sets are presented in Table~\ref{tab:toy_mi}, showing that while all lower bounds perform well, the JSD lower bound results in the highest MI lower bound estimate. We note that, even though all lower bounds had the same neural network and design initialisations, the InfoNCE lower bound in particular would sometimes find slightly different (locally) optimal designs; we show an example of this in the supplementary materials. 

\begin{figure}[!t]
    \centering
    \includegraphics[width=\linewidth]{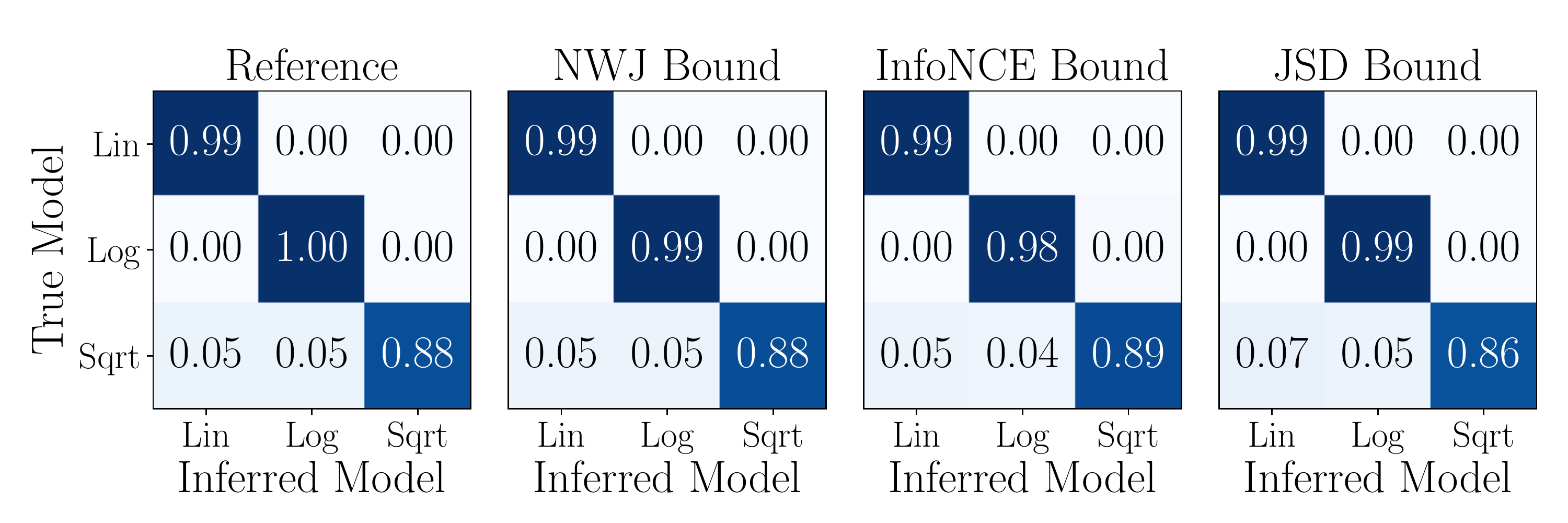}
    \vspace{-0.5cm}
    \caption[]{MD posterior results for the set of toy models. Shown are average posterior probabilities for different ground truth models (one per row) for the NWJ, InfoNCE and JSD lower  bounds.}
    \label{fig:toy_md_post}
\end{figure}

Similar to the PE task, we can use the trained neural networks to find posterior distributions over the model indicator $m$, which express our posterior belief that a given model has generated the observed `real-world' data $\ybf^\ast$. We generate three different sets of `real-world' data, one for each of the three models being the underlying true model. We also use the same model parameter ground truth $\thetab_{m, \text{true}} = (2, 3)$ for each model. Figure~\ref{fig:toy_md_post} shows average posterior distributions for all lower bounds, including an average reference posterior distribution. All methods result in extremely good model recovery, meaning that they can, on average, confidently identify from which model the data was generated.
The difference in performance between the different lower bounds becomes more pronounced when considering the average KL-Divergence between estimated and reference posteriors, as shown in Table~\ref{tab:toy_kl}. The average KL-Divergences show that the JSD lower bound results in posterior distributions that are slightly closer to the reference than NWJ and InfoNCE.

\subsubsection{Joint MD/PE}

Here we consider the joint task of model discrimination and parameter estimation (MD/PE), i.e.~a combination of the previous two tasks. This means that our variable of interest is now the set of model indicator and corresponding model parameters, i.e.~$\vintbf = (\thetab_m, m)$.
The MD/PE results are similar to previous results seen for the separate parameter estimation and model discrimination tasks. As such, we only show the relevant training curves and posterior distributions in the supplementary materials. The training curves look similar for all lower bounds and the optimal designs $\dbf^\ast$ were also approximately the same. The elements of $\dbf^\ast$ were clustered around $d=-2$, $d=0$ and $d=2$, similar to those found in the MD task. In the MD/PE column in Table~\ref{tab:toy_mi} we present final MI estimates for each lower bound, evaluated on several validation data sets. The JSD lower bound performs significantly better than the NWJ and InfoNCE lower bounds, i.e.~it is closer to the reference value. Similarly, Table~\ref{tab:toy_kl} shows that the posteriors estimated via JSD have, on average, a lower KL-Divergence to reference posteriors, compared to the other lower bounds. Furthermore, the NWJ lower bound results in an extremely large average KL-Divergence, which is in part due to a poor parameter estimation for the logarithmic model (see the supplementary materials for a plot).

\subsubsection{Improving Future Predictions for the Linear Model}

Finally, we consider the task of improving future predictions (FP), as discussed in Section~\ref{sec:background}, for the linear toy model. The aim here is to find current optimal designs $\dbf^\ast$, and gather corresponding data $\ybf^\ast$, that allow us to maximally improve our predictions of future data $y_T$ gathered at a (fixed and known) future design $d_T$. Our variable of interest is thus the future data $y_T$. As before, we construct a 10-dimensional design vector $\dbf$ that has elements restricted to the domain $d \in [-2, 2]$, with a corresponding 10-dimensional data vector $\ybf$. For simplicity, we assume that our future design is one-dimensional and fixed at $d_T = 4$, which is outside the domain of our current design vector. This emulates a setting where, for instance, we know that we will be able to make measurements in a specific geographical location of interest but currently only have access to a different, limited region, allowing us to improve our prediction at that future measurement location. This setting naturally assumes that the implicit model extrapolates well to future designs $d_T$ when they are outside the current design domain.

\pagebreak

The training curves for the NWJ, InfoNCE and JSD lower bound again show a very similar convergence behaviour and yield the same optimal designs. As such, we only show the training curves in the supplementary materials. The optimal designs $\dbf^\ast$ have elements that are clustered equally at the boundaries, which is exactly the same as for the parameter estimation task. These designs are useful for improving future predictions, as the parameter estimation is best (see the PE section) and the absolute values are close to that of the future design. Table~\ref{tab:toy_mi} shows that the final MI estimates for each lower bound are similar and nearly match the reference value, with the JSD lower bound yielding a slightly closer estimate.


\begin{figure}[!t]
    \centering
    \includegraphics[width=\linewidth]{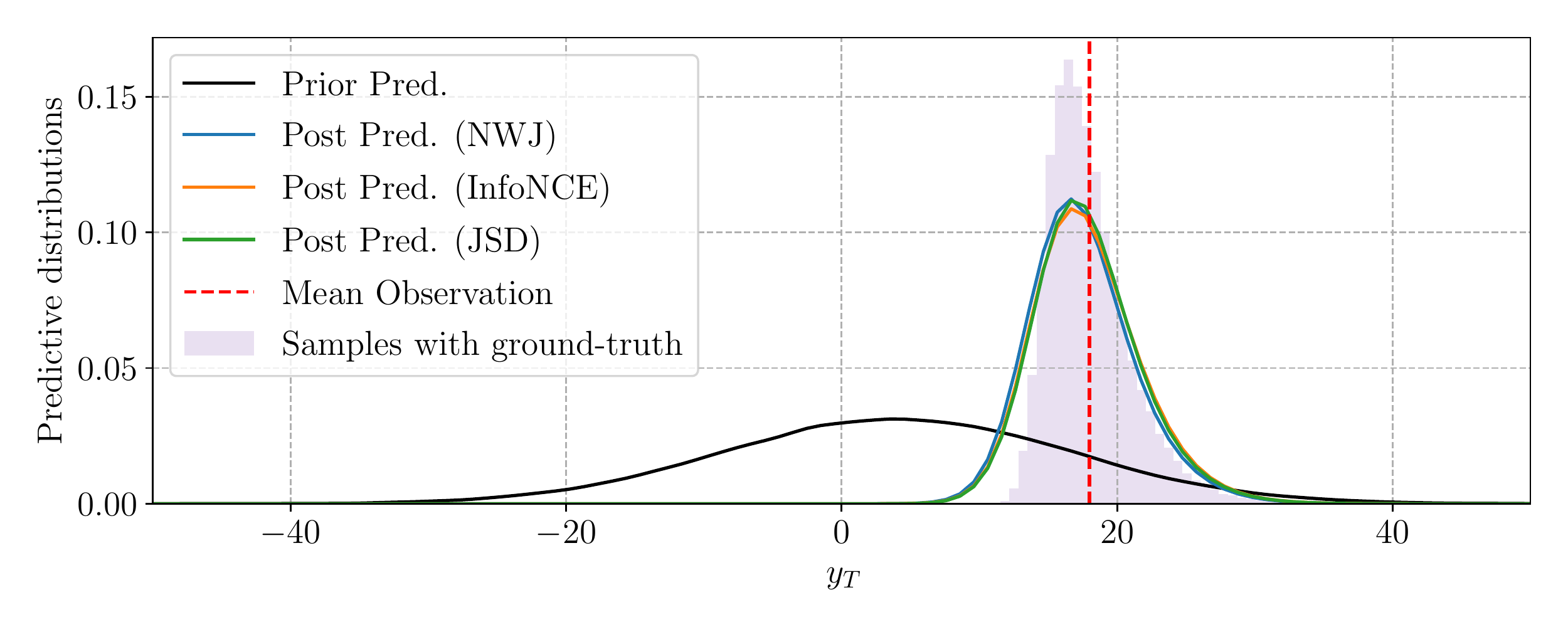}
    \vspace{-0.7cm}
    \caption[]{FP results for the linear toy model. Shown are average posterior predictive distributions for the NWJ, InfoNCE and JSD lower bounds. Shown are also the prior predictive and a histogram of `real-world' samples at the future design.}
    \label{fig:toy_fp_post}
\end{figure}

For the FP task, we can use the trained neural networks to obtain a posterior predictive distribution $p(y_T|d_T, \ybf^\ast, \dbf^\ast)$ of our variable of interest $y_T|d_T$. To generate `real-world' data $\ybf^\ast$ at $\dbf^\ast$ we use ground-truth values of $\thetab_{\text{true}} = (2, 3)$ for the offset and slope, respectively. In Figure~\ref{fig:toy_fp_post} we show the average posterior predictive distributions, averaged over $5{,}000$ different $\ybf^\ast$, for all lower bounds, including the prior predictive distribution and a histogram showing samples of the data-generating distribution of $y_T$ at $d_T$ with $\thetab_{\text{true}}$ (which is not a predictive distribution). The posterior predictive distributions for all lower bounds are similar to each other and all have a mode that matches that of the histogram of `real-world' data at $d_T$. The differences between the lower bounds becomes more noticeable when looking at the average KL-Divergence values shown in Table~\ref{tab:toy_kl}. As for the other scientific tasks, the JSD lower bound results in posteriors that have the lowest average KL-Divergence to reference posteriors.

\subsection{SDE Epidemiology Model}

\begin{figure}[!t]
    \centering
    \includegraphics[width=1\linewidth]{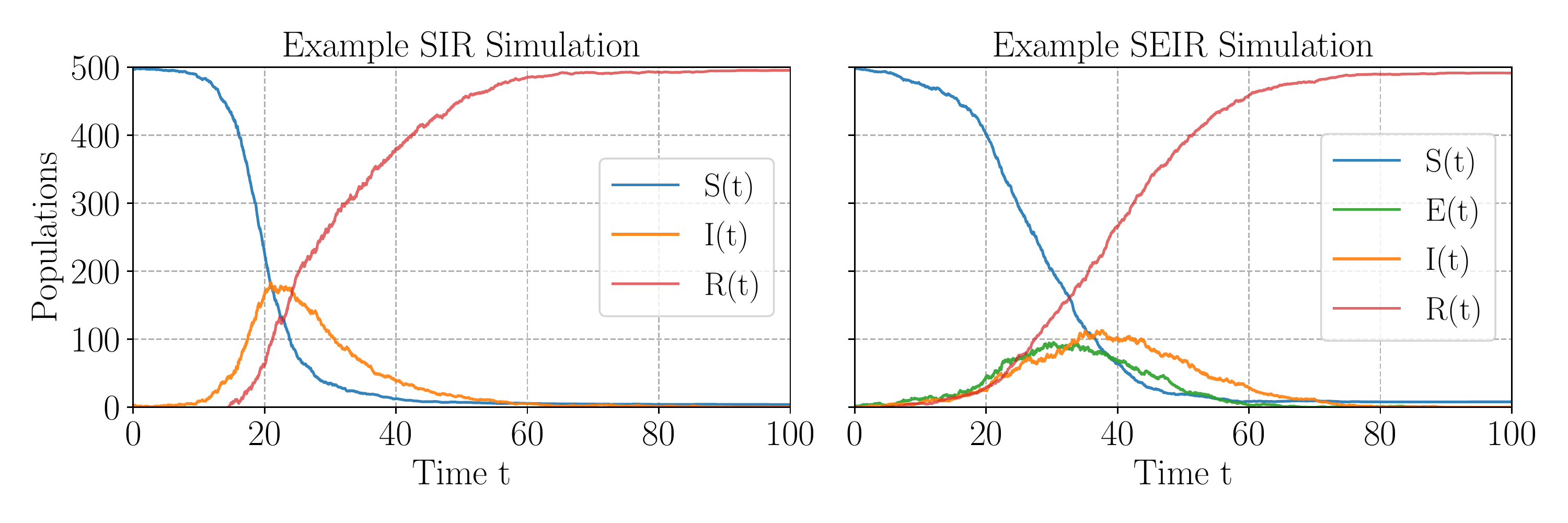}
    \vspace{-0.7cm}
    \caption[]{Example simulations of the SDE-based SIR (left) and SEIR (right) models.}
    \label{fig:examples}
\end{figure}

We here consider the spread of a disease within a population of $N$ individuals, modelled by stochastic versions of the well-known SIR~\citep{Allen2008} and SEIR~\citep{Lekone2006} models. In the SIR model, indicated by $m=1$, individuals start in a susceptible state $S(t)$ and can then move to an infectious state $I(t)$ with an infection rate of $\beta$. These infectious individuals then move to a recovered state $R(t)$ with a recovery rate of $\gamma$, after which they can no longer be infected. The SIR model, governed by the state changes $S(t) \rightarrow I(t) \rightarrow R(t)$, thus has two model parameters $\thetab_1 = (\beta, \gamma)$. In the SEIR model, indicated by $m=2$, susceptibles first move to an additional exposed state $E(t)$, where individuals are infected but not yet infectious. Afterwards, they move to the infectious state $I(t)$ with a rate of $\sigma$. The SEIR model ($m=2$), governed by $S(t) \rightarrow E(t) \rightarrow I(t) \rightarrow R(t)$, thus has three model parameters $\thetab_2 = (\beta, \sigma, \gamma)$. We further make the common assumption that the total population size $N$ stays constant. 

The stochastic versions of these epidemiological processes are usually defined by a continuous-time Markov chain (CTMC), from which we can sample via the Gillespie algorithm~\citep[see][]{Allen2017}. However, this generally yields discrete population states that have undefined gradients.
In order to test our gradient-based algorithm, we thus resort to an alternative simulation algorithm that uses stochastic differential equations (SDEs), where gradients can be approximated. Figure~\ref{fig:examples} shows example simulations of the SDE-based SIR and SEIR models, generated according to the method below.

We first define population vectors $\mathbf{X}_1 = (S_1(t), I_1(t))^\top$ for the SIR model and $\mathbf{X}_2 = (S_2(t), E_2(t), I_2(t))^\top$ for the SEIR model. We can effectively ignore the population of recovered because the total population is fixed.\footnote{Allowing us to, e.g., compute~$R_2(t) = N - S_2(t) - E_2(t) - I_2(t)$ for the SEIR model.} The system of Itô SDEs for the above epidemiological processes is
\begin{equation} \label{eq:sde}
\mathrm{d}\mathbf{X}_m(t) = \mathbf{f}_m(\mathbf{X}_m(t)) \mathrm{d}t + \mathbf{G}_m(\mathbf{X}_m(t))\mathrm{d}\mathbf{W}(t),
\end{equation}
where $m$ is a model indicator, $\mathbf{f}_m$ is called the drift vector, $\mathbf{G}_m$ is called the diffusion matrix, and $\mathbf{W}(t)$ is a vector of independent Wiener processes (also called Brownian motion). The drift vector and diffusion matrix for both models are derived in detail in the supplementary materials.
%
%
Sample paths of the above SDEs can be simulated by means of the Euler-Maruyama algorithm~\citep[see][]{Allen2017}, which is based on finite differences and discretised Wiener processes. In our experiments we use the $\texttt{torchsde}$ package~\citep{Li2020} to do this efficiently.

Moreover, we assume that we only have access to discrete, noisy estimates of the number of infectious individuals $I(t)$, as other population states might be difficult to measure in reality. We thus sample a single noisy measurement of $I_m(t)$ by
\begin{equation} \label{eq:obs}
    y | t, m, \thetab_m \sim \text{Poisson}(y; \phi I_m(t)),
\end{equation}
where $\phi$ is assumed to be known and set to $0.95$ throughout our experiments.\footnote{While we have not done so, our method would allow us to treat $\phi$ as a model parameter as well.} This corresponds to the setting of a known observational model described by Equation~\ref{eq:exp_latent}.


The design variable in our experiments is the measurement time $t$ in Equation~\ref{eq:obs}. We generally wish to take several measurements at once and hence define a design vector $\dbf = (t_1, \dots, t_D)$ with corresponding observations $\ybf = (y_1, \dots, y_D)$, where $D$ is the number of measurements.
In our experiments, we use $N=500$ with initial conditions of $\mathbf{X}_1(t=0) = (498, 2, 0)^\top$ for the SIR model and $\mathbf{X}_2(t=0) = (498, 0, 2, 0)^\top$ for the SEIR model. We also use a discretised design space of $t \in [0, 100]$ with resolution of $\Delta t = 10^{-2}$. For the model parameter priors we use $p(\beta) = \text{Lognorm}(0.50, 0.50^2)$, $p(\sigma) = \text{Lognorm}(0.20, 0.50^2)$ and $p(\gamma) = \text{Lognorm}(0.10, 0.50^2)$. Further information about neural network architectures, hyper-parameters and experimental details are given in the supplementary materials.

\subsubsection{Parameter Estimation for the SIR Model}

\begin{figure}[!t]
    \centering
    \includegraphics[width=\linewidth]{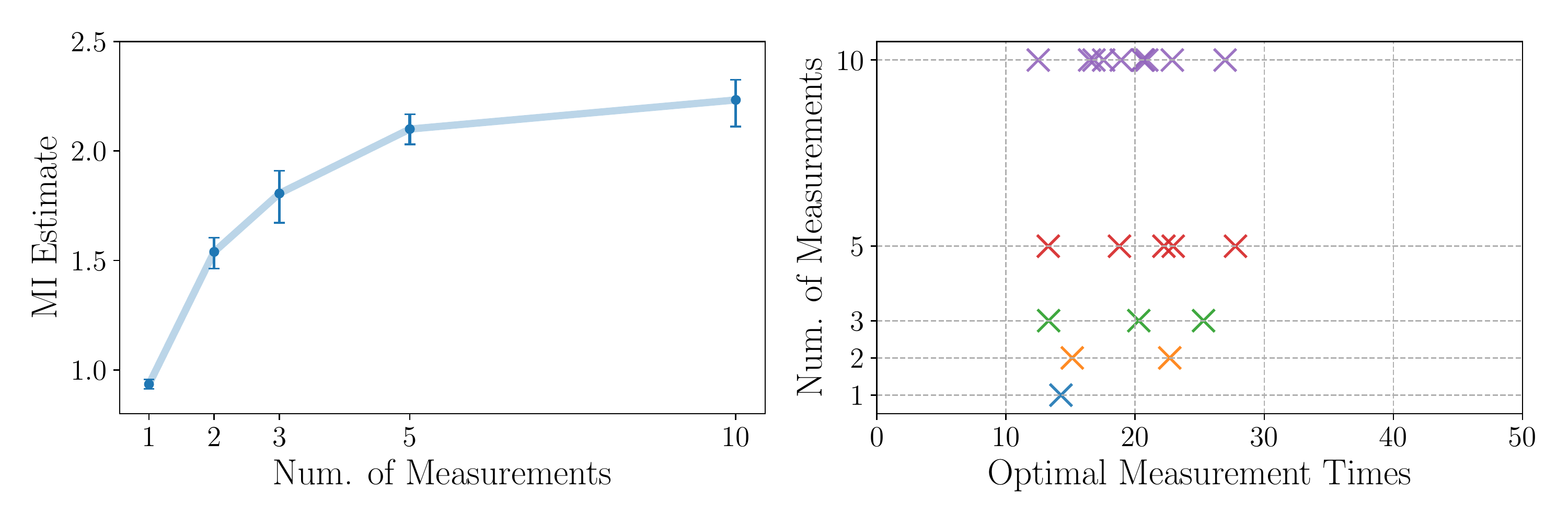}
    \vspace{-0.7cm}
    \caption[]{PE results for the SDE-based SIR model with different experimental budgets. The left shows validation MI estimates for different number of measurements, averaged over several validation data sets. The right plot shows the corresponding optimal measurement times.}
    \label{fig:sde_pe_results}
\end{figure}

First, we consider the task of parameter estimation (PE) for the SIR model, where the aim is to estimate its model parameters $\thetab_1 = (\beta, \gamma)$. We here use the JSD lower bound as an objective function and test different experimental budgets of up to $10$ measurements. 
Figure~\ref{fig:sde_pe_results} summarises the results for settings with different budgets. The left plot shows validation estimates of the maximum MI at the respective optimal designs (shown in the right plot), obtained through our proposed method, for varying numbers of measurements $D$. The optimal designs $\dbf^\ast$ were all in the region between $t=10$ and $t=40$, with a slightly different spreading for different $D$. Intuitively, this region might be useful because that is where the signal-to-noise ratio of the SIR observations is highest.\footnote{See the supplementary materials for a plot showing this.} As before in the PE task for the linear toy model, this means that we are able to better measure the effect of model parameters on observations. The maximum MI in the left of Figure~\ref{fig:sde_pe_results} starts to saturate at $D=10$ measurements, suggesting that more measurements may not improve parameter estimation by much. 


\begin{figure}[!t]
    \centering
    \includegraphics[width=\linewidth]{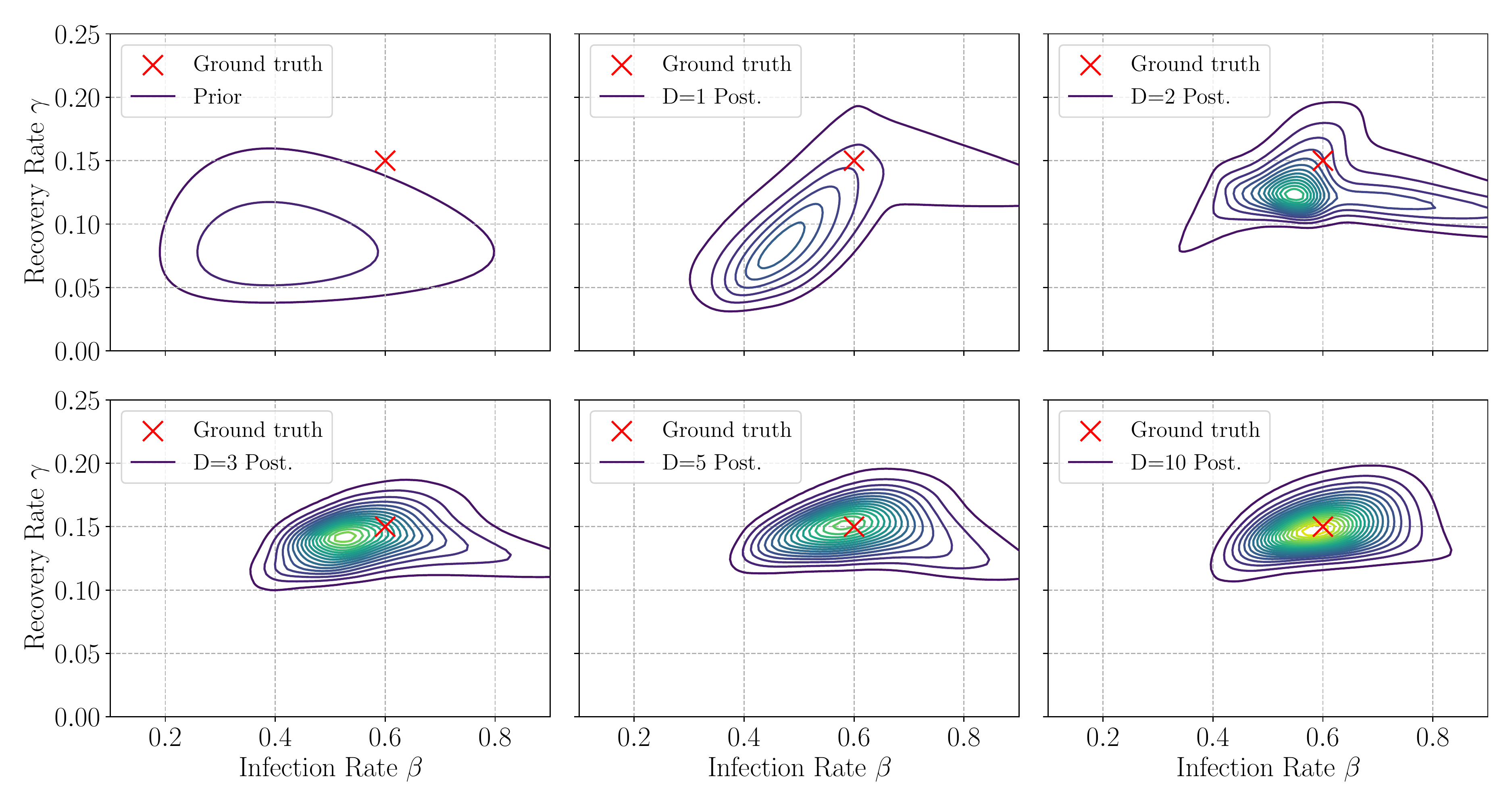}
    \vspace{-0.7cm}
    \caption[]{PE posterior distributions for the SDE-based SIR model with different experimental budgets. The top left shows the prior distribution, while the other plots show posteriors for an increasing number of measurements. The posteriors are averaged over $5{,}000$ `real-world' observations $\ybf^\ast$ generated with ground-truth $(0.60, 0.15)$.}
    \label{fig:sde_pe_postsumm}
\end{figure}

We can estimate posterior distributions of the model parameters $(\beta, \gamma)$ by means of the trained neural network (see Section~\ref{sec:methods}). To do so, we generate `real-world' observations $\ybf^\ast$ at $\dbf^\ast$ using the ground truth $\thetab_{\text{true}} = (0.60, 0.15)$.
Figure~\ref{fig:sde_pe_postsumm} shows the prior distribution (top left) and average posterior distributions for different number of measurements, where the average is taken over $5{,}000$ observations $\ybf^\ast$. As we increase the number of measurements, the average posterior naturally becomes more peaked and moves closer to the ground truth parameters (indicated by a red cross). At $D=10$, the ground truth is well-captured by the average posterior, with an average MAP estimate, taken over all `real-world' observations, of $\widehat{\thetab}_{\text{MAP}} = (0.59, 0.07) \pm (0.15, 0.01)$, where the errors indicate one standard deviation. All training curves, posterior distributions and optimal designs are shown in the supplementary materials.


\subsubsection{Model Discrimination}

Next, we consider the task of model discrimination (MD) between the SIR and SEIR model. We again use the JSD lower bound as a the utility function and test different experimental budgets of up to $10$ measurements. The results for this setting are summarised in Figure~\ref{fig:sde_md_results}. The left plot shows optimal designs for varying numbers of measurements $D$, obtained through our proposed method. The optimal designs have elements that are clustered around smaller measurement times between $t=15$ and $t=40$, similar to the optimal designs for the PE task in Figure~\ref{fig:sde_pe_results}.
These measurement times might be useful because that is where the average signal-to-noise ratio is highest for both models and the corresponding data distributions are significantly different.\footnote{See the supplementary material for plots showing this.} This means that we can gather (relatively) low-noise data that can be related to the corresponding models more effectively. We show corresponding validation estimates of the MI as a function of $D$ in the supplementary materials. Similar to the PE task, the MI tends to increase as the number of measurements are increased, although the change is much more subtle for the MD task.

\begin{figure}[!t]
    \centering
    \includegraphics[width=\linewidth]{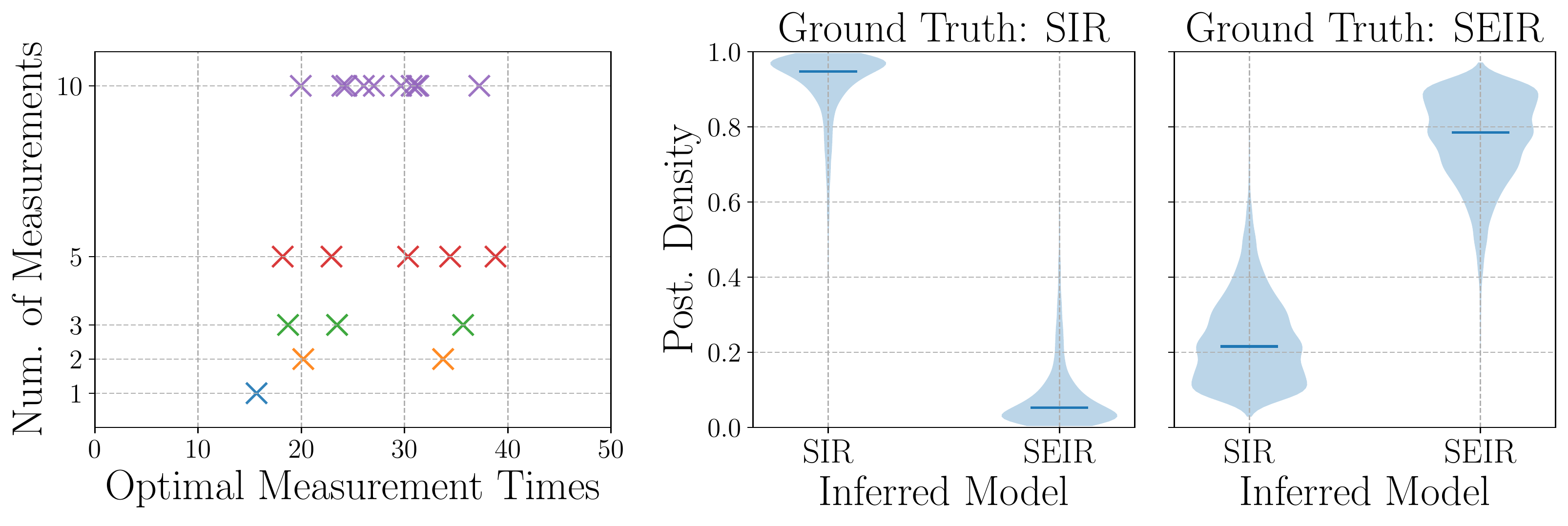}
    \vspace{-0.7cm}
    \caption[]{MD results for the SDE-based epidemiology models. The left plot shows optimal experimental designs for varying number of measurements. The other plots show distributions of posterior densities for $D=10$ measurements, given that the SIR model (middle plot) or the SEIR model (right plot) is the ground truth.}
    \label{fig:sde_md_results}
\end{figure}

For the case of $D=10$ measurements, the middle and right plots in Figure~\ref{fig:sde_md_results} show distributions of posterior densities $p(m|\dbf^\ast, \ybf^\ast)$ estimated using several $\ybf^\ast$, with the middle plot assuming that SIR is the ground-truth model and the right plot assuming that SEIR is the ground-truth model. We can recover the ground-truth model effectively in both cases, with corresponding F1-scores of $0.996$ and $0.983$. However, the model recovery is slightly worse when the SEIR model is the ground truth, which we discuss further in the supplementary materials. 


\section{Conclusions} \label{sec:concl}
In this work we have introduced a framework for Bayesian experimental design with implicit models. Our approach finds optimal experimental designs by maximising general lower bounds on mutual information that are parametrised by neural networks. We have derived gradients of prominent lower bounds with respect to the experimental designs, allowing us to perform stochastic gradient-ascent. In doing so, we have provided a methodology to derive gradients of general lower bounds on mutual information, which may be applied to other lower bounds that are introduced in the future. Furthermore, our framework yields an amortised posterior distribution as a by-product, as long as the utilised lower bound facilitates this, which is generally a challenging problem for implicit models and has an entire research field dedicated to it~\citep[see][for recent overviews]{Lintusaari2017, Sisson2018, Cranmer2020}.

By means of a set of intractable toy models, we have provided a thorough experimental comparison of several prominent lower bounds for a variety of scientific tasks, showcasing the versatility of our proposed framework. We have also applied our method to a challenging set of implicit models from epidemiology, where the responses are discrete observations of the solutions to stochastic differential equations. Our approach allowed us to efficiently obtain optimal designs, leading to suitable posterior distributions that capture ground-truths well. Throughout our experiments we have considered popular scientific tasks such as parameter estimation, model discrimination and improving future predictions. Importantly, we believe that our framework would also be able to deal with other scientific tasks that are formulated using mutual information.

Our approach requires us to have access to gradients of the sampling path with respect to experimental designs, or that we can compute them by means of automatic differentiation. There may be cases where these gradients are unknown or undefined, e.g.~when data is discrete or categorical. We note that our framework can handle those cases when the corresponding observational process has an analytically tractable model, but currently not otherwise.~\citet{Kleinegesse2020b} have shown that it is still possible to use gradient-free optimisation techniques to find optimal designs, following the method explained in Section~\ref{sec:methods}. However, these approaches do not scale well with design dimensions, unlike gradient-based approaches~\citep[see][]{Spall2003}. It would be interesting to explore how our method can be scalably adapted to these situations.

In related work,~\citet{Foster2019} perform gradient-based Bayesian experimental design for implicit models by using variational approximations to likelihoods or posteriors. This allows them to use score-function estimators, as opposed to pathwise gradient estimators. Recently,~\citet{Zhang2021} have used evolutionary strategies to approximate gradients of the SMILE lower bound in the context of Bayesian experimental design. There are several other interesting methods that allow us to approximate gradients during the optimisation procedure, e.g.~finite-difference or simultaneous perturbation stochastic approximations~\citep{Spall2003}, that have not been investigated yet. As such, investigating avenues that mitigate or relax the need to have access to gradients of the sampling path
may increase the applicability of our method further.

We have applied our framework to the setting of static Bayesian experimental design, where we wish to determine a set of optimal designs prior to actually performing the experiment. Sequential Bayesian experimental design, however, advocates that we should update our belief of the variable of interest as we perform the experiment~\citep[see e.g.][]{Kleinegesse2020a}. It would be interesting to apply our framework to this setting as well. In particular, ~\citet{Foster2021} have recently introduced an approach that yields a design network which directly outputs optimal designs and, as such, allows for amortised sequential Bayesian experimental design, greatly increasing practicability. While their method is designed for explicit models that have known likelihoods, we believe that it would be interesting to apply our method of maximising parametrised lower bounds to their method as well.

Furthermore, in this work we have mostly glanced over the neural network architecture selection. However, we believe that there is a lot still to be explored that may help during the training procedure, e.g.~residual connections, separable critics, pooling, convolutions, and other techniques that leverage known structures. In particular, it may be useful to directly incorporate a dependency on the designs into the neural network, e.g.~by using the current design as an input as well. This may strengthen the extrapolation of the neural network, improving training speed and robustness. We also believe that the interplay between the number of dimensions of the variable of interest $\vintbf$ and the data variable $\ybf$ is highly important. If one is much larger than the other, the neural network $\NN$ may have more difficulties learning the appropriate density ratio. It may be useful to use a separable critic, or automatically learn summary statistics that reduce the dimensionality while retaining all the necessary information as e.g.\ recently done by~\citet{Chen2021} in the field of likelihood-free inference.


\clearpage


\appendix

\section{MI Lower bounds}

\subsection{Relationship between the JSD lower bound and LFIRE}

Likelihood-free inference by ratio estimation~\citep[LFIRE,][]{Thomas2020} aims to approximate the density ratio $r(x, \theta)$ of the data-generating distribution $p(x|\theta)$ and the marginal distribution $p(x)$ for implicit models, where $p(x|\theta)$ is intractable but sampling from it is still possible. Using a known prior distribution $p(\theta)$, this learned density ratio can then be used to compute the posterior distribution $p(\theta|x)$. Importantly, LFIRE only requires samples $x^{\theta}_i \sim p(x|\theta)$ from the data-generating distribution and samples $x^{m}_i \sim p(x)$ from the marginal distribution.

This likelihood-free inference method works by formulating a classification problem between data sampled from $p(x|\theta)$ and $p(x)$, and then solving this via (non-linear) logistic regression. Using their notation, the loss function that they minimise is
\begin{equation} \label{eq:lfire_loss}
    \mathcal{J}(h, \theta) = \frac{1}{n_{\theta}+n_m}
    \left\{
    \sum_{i=1}^{n_{\theta}}
    \log\left[1+\nu e^{-h(x^{\theta}_i)}\right]
    + \sum_{i=1}^{n_m}
    \log\left[1+\frac{1}{\nu}e^{h(x^m_i)}\right]
    \right\},
\end{equation}
where $n_{\theta}$ and $n_m$ are the number of samples from $p(x|\theta)$ and $p(x)$, respectively, $\nu = n_{\theta} / n_m$ is used to correct unequal class sizes, and $h$ is a non-linear, parametrised function. The authors prove that for large $n_{\theta}$ and $n_m$, the function $h^{\ast}$ that minimises $\mathcal{J}(h, \theta)$ is given by the desired log-ratio, i.e.~$h^{\ast}(x, \theta) = \log r(x, \theta)$. Importantly, this result holds for any a fixed $\theta$, so that the result still holds if we consider the objective $\mathcal{J}(h, \theta)$ averaged over $\theta$.

Recall that the JSD lower bound presented in the main text is given by
\begin{equation} \label{eq:jsd}
\mathcal{L}_{\text{JSD}}(\psib, \dbf) \equiv  \Ej \left[ -\text{sp}(-\NN) \right] - \Em \left[ \text{sp}(\NN) \right],
\end{equation}
where $\text{sp}(z) = \log(1 + e^z)$ is the softplus function and $\NN$ is a neural network with parameters $\psib$. The variables $\ybf, \vintbf$ and $\dbf$ are the data, variable of interest and experimental designs, respectively. Our BED framework works by maximising Equation~\ref{eq:jsd} with respect to $\psib$ and $\dbf$ to find optimal designs $\dbf$. Let us here focus on the specific case of fixed $\dbf$, i.e.~only $\psib$ is updated. We here aim to show the strong relationship between the JSD lower bound and the objective used in LFIRE.

First, let us make the notational changes $\vintbf \rightarrow \theta$, $\ybf | \dbf \rightarrow x$ and $T_{\psib} \rightarrow h$, in order to match the notation of~\citet{Thomas2020}. Equation~\ref{eq:jsd} then becomes
\begin{align} \label{eq:jsd_change}
\mathcal{L}_{\text{JSD}}(h) 
&=  \mathbb{E}_{p(x, \theta)} \left[ -\text{sp}(-h(x, \theta) \right] - \mathbb{E}_{p(x)p(\theta)} \left[ \text{sp}(h(x, \theta)) \right] \\
&= \mathbb{E}_{p(x, \theta)} \left[ -\log\left(1 + e^{-h(x, \theta)}\right) \right] - \mathbb{E}_{p(x)p(\theta)} \left[ \log\left(1 + e^{h(x, \theta)}\right) \right].
\end{align}
Writing $p(x, \theta) = p(x|\theta)p(\theta)$ we can move the expectation $\mathbb{E}_{p(\theta)}$ outside, as it is present in both terms in the above equation. The resulting JSD lower bound can then be seen as an expectation of the LFIRE objective in Equation~\ref{eq:lfire_loss}, with equal class sizes $n_{\theta}=n_m$ and large number of samples $n_m$, $n_{\theta} \rightarrow \infty$, i.e.
\begin{align} \label{eq:jsd_change}
\mathcal{L}_{\text{JSD}}(h) 
&= -\mathbb{E}_{p(\theta)} \left\{ \mathbb{E}_{p(x \mid \theta)} \left[ \log\left(1 + e^{-h(x, \theta)}\right) \right] + \mathbb{E}_{p(x)} \left[ \log\left(1 + e^{h(x, \theta)}\right) \right] \right\} \\
&= -\mathbb{E}_{p(\theta)} \left\{ \mathcal{J}(h, \theta) \right\}. \label{eq:lfire_jsd_equiv}
\end{align}

Recall that LFIRE is used to learn the density ratio $r(x, \theta)$ for a particular value of $\theta$ (with an amortisation in $x$). As explained in the main text, a neural network trained with the JSD lower bound can be used to directly compute the log density ratio $\log r(x, \theta)$, but with amortisation in $\theta$ as well as $x$. This fact and Equation~\ref{eq:lfire_jsd_equiv} imply that maximising the JSD lower bound can be seen as minimising the LFIRE objective in an amortised fashion. While LFIRE and the JSD lower bound have existed in literature for quite some time, we believe that this strong relationship between them has previously not been recorded. Moreover, LFIRE has been used before in the context of Bayesian experimental design for implicit models~\citep{Kleinegesse2019, Kleinegesse2020a}.

\subsection{Derivation of lower bound gradients}

Using the methodology presented in Section 3.3 of the main text, we here provide detailed derivations of the lower bound gradients shown in Table 2 of the same section. Recall that mutual information lower bounds in literature generally involve expectations $\Ej \left[ f(\NN) \right]$ over the joint and/or expectations $\Em \left[ g(\NN) \right]$ over the marginal distribution, where $f:\mathbb{R}\rightarrow \mathbb{R}$ and $g:\mathbb{R}\rightarrow\mathbb{R}$ are non-linear, differentiable functions. As shown in Equations 3.10 and 3.11 in the main text, we can compute gradients of these expectations by means of pathwise gradient estimators (i.e.~the reparametrisation trick). 

In the interest of space, let us define $T = T_{\bm{\psi}}(\bm{\vint}, \mathbf{h}(\bm{\epsilon}; \vintbf, \dbf))$, $\widetilde{T} = T_{\bm{\psi}}(\bm{\vint}, \mathbf{h}(\bm{\epsilon}; \widetilde{\vintbf}, \dbf))$, as well as $P=p(\vintbf)p(\bm{\epsilon})$ and $Q=p(\vintbf)p(\widetilde{\vintbf})p(\bm{\epsilon})$, where $p(\widetilde{\vintbf})$ is the same as $p(\vintbf)$. Importantly, $P$ and $Q$ do not depend on $\dbf$. Our methodology in the main text requires us to first compute $f^{\prime}(T) = \partial f(T) / \partial T$ and $g^{\prime}(\widetilde{T}) = \partial g(\widetilde{T}) / \partial \widetilde{T}$ in order to compute gradients of lower bounds with respect to designs. Below we show step-by-step derivations of the gradients for all lower bounds shown in Table 2 of the main text.


\paragraph{NWJ} Defined in Equation 3.2 in the main text, the NWJ lower bound involves the functions $f(T) = T$ and $g(\widetilde{T}) = e^{\widetilde{T}-1}$, which yield the corresponding gradients $\nabla_T f(T) = 1$ and $\nabla_{\widetilde{T}} g(\widetilde{T}) = e^{\widetilde{T}-1}$. The desired gradients of the NWJ lower bound with respect to designs $\dbf$ are then computed as follows,
\begin{align}
\nabla_{\dbf} \mathcal{L}_{\text{NWJ}}(\psib, \dbf) 
&= \nabla_{\dbf} \left\{ \mathbb{E}_P \left[ f(T) \right] - \mathbb{E}_Q \left[ g(\widetilde{T}) \right] \right\} \\
&= \mathbb{E}_P \left[ \nabla_{\dbf} f(T) \right] - \mathbb{E}_Q \left[ \nabla_{\dbf} g(\widetilde{T}) \right] \\
&= \mathbb{E}_P \left[ \nabla_{T} f(T) \nabla_{\dbf} T \right] - \mathbb{E}_Q \left[ \nabla_{\widetilde{T}} g(\widetilde{T}) \nabla_{\dbf} \widetilde{T} \right] \\
&= \mathbb{E}_P \left[\nabla_{\dbf} T \right] - \mathbb{E}_Q \left[ e^{\widetilde{T}-1} \nabla_{\dbf} \widetilde{T} \right],
\end{align}
where $\nabla_{\dbf} T$ and $\nabla_{\dbf} \widetilde{T}$ are given by Equation 3.12 in the main text.

\paragraph{InfoNCE} This lower bound, defined in Equation 3.3 in the main text, only involves an expectation over the joint distribution and not the marginal distribution, i.e.~$g(T) = 0$. In addition to the previous shorthands, let us also define $T_{ij}=T_{\psib}(\vintbf_j,h(\bm{\epsilon}_i;\vintbf_i,\dbf))$. In this particular case, it is simpler to directly start with the gradient with respect to $\dbf$, yielding
\begin{align}
\nabla_{\dbf} \mathcal{L}_{\text{NCE}}(\psib, \dbf)
&= \nabla_{\dbf} \mathbb{E}_{P^K} \left[  \frac{1}{K} \sum_{i=1}^K \log{\frac{e^{T_{ii}}}{\frac{1}{K} \sum_{j=1}^K e^{T_{ij}} }} \right] \\
&= \mathbb{E}_{P^K} \left[ \frac{1}{K} \sum_{i=1}^K \nabla_{\dbf} \log{\frac{e^{T_{ii}}}{\frac{1}{K} \sum_{j=1}^K e^{T_{ij}} }} \right] \\
&= \mathbb{E}_{P^K} \left[ \frac{1}{K} \sum_{i=1}^K \nabla_{\dbf} T_{ii} - \nabla_{\dbf} \log{\sum_{j=1}^K e^{T_{ij}} } + \nabla_{\dbf} \log{K} \right] \\
&= \mathbb{E}_{P^K} \left[ \frac{1}{K} \sum_{i=1}^K \nabla_{\dbf} T_{ii} - \nabla_{\dbf} \log{\sum_{j=1}^K e^{T_{ij}} } \right] \\
&= \mathbb{E}_{P^K} \left[ \frac{1}{K} \sum_{i=1}^K \nabla_{\dbf} T_{ii} - \frac{\sum_{j=1}^K e^{T_{ij}} \nabla_{\dbf} T_{ij} }{\sum_{j=1}^K e^{T_{ij}}} \right] \\
&= \mathbb{E}_{P^K} \left[ \frac{1}{K} \sum_{i=1}^K \frac{\sum_{j=1}^K e^{T_{ij}} (\nabla_{\dbf} T_{ii} - \nabla_{\dbf} T_{ij})}{\sum_{j=1}^K e^{T_{ij}}}\right], \label{eq:infonce_last}
\end{align}
where $\nabla_{\dbf} T_{ii}$ and $\nabla_{\dbf} T_{ij}$ are again given by Equation 3.12 in the main text. As explained in the main text, the above formulations yields an optimal critic $\NNopt = \log{\likeint} + c(\ybf|\dbf)$, where $c(\ybf|\dbf)$ is an indeterminate function. 


\paragraph{JSD} The JSD lower bound is defined in Equation 3.4 in the main text and involves expectations of the non-linear functions $f(T) = -\text{sp}(-T)$ and $g(\widetilde{T}) = \text{sp}(\widetilde{T})$, where $\text{sp}(z) = \log\left(1 + e^{z}\right)$ is the softplus function. The derivative of the softplus function is simply the logistic sigmoid function, i.e.~$\nabla_z \text{sp}(z) = 1 / (1 + e^{-z}) \equiv \sigma(z)$, resulting in the gradients $\nabla_T f(T) = \sigma(-T)$ and $\nabla_{\widetilde{T}} g(\widetilde{T}) = \sigma(\widetilde{T})$. This allows us to compute the gradients of the JSD lower bound with respect to experimental designs,
\begin{align}
\nabla_{\dbf} \mathcal{L}_{\text{JSD}}(\psib, \dbf) 
&= \nabla_{\dbf} \left\{ \mathbb{E}_P \left[ f(T) \right] - \mathbb{E}_Q \left[ g(\widetilde{T}) \right] \right\} \\
&= \mathbb{E}_P \left[ \nabla_{\dbf} f(T) \right] - \mathbb{E}_Q \left[ \nabla_{\dbf} g(\widetilde{T}) \right] \\
&= \mathbb{E}_P \left[ \nabla_{T} f(T) \nabla_{\dbf} T \right] - \mathbb{E}_Q \left[ \nabla_{\widetilde{T}} g(\widetilde{T}) \nabla_{\dbf} \widetilde{T} \right] \\
&= \mathbb{E}_P \left[\sigma(-T) \nabla_{\dbf} T \right] - \mathbb{E}_Q \left[ \sigma(\widetilde{T}) \nabla_{\dbf} \widetilde{T} \right],
\end{align}
where $\nabla_{\dbf} T$ and $\nabla_{\dbf} \widetilde{T}$ are again computed using Equation 3.12 in the main text.

\subsection{Gradient estimator for analytically tractable observation models}

We here provide the gradient estimator of mutual information lower bounds in situations where it is possible to leverage an analytically tractable observation model, as discussed in Section 3.4 of the main text. Recall that we assume that we can separate the data generation into an observation process with a model $p(\ybf|\vintbf, \dbf, \bm{z})$ that is known analytically in closed form and a differentiable, latent process $p(\bm{z}|\vintbf, \dbf)$. We can reparametrise the latent variable using its sampling path, i.e.~$\bm{z} = \bm{h}(\bm{\epsilon}; \vintbf, \dbf)$, where the noise random variable $\bm{\epsilon}$ defines the sampling path. This allows us to use score-function estimators to compute the required gradients.

The score function estimator works by re-writing the derivative of $p(\ybf|\vintbf, \dbf, \bm{z})$ using the definition of the logarithmic derivative, i.e.
\begin{equation}
    \nabla_{\dbf} p(\ybf|\vintbf, \dbf, \bm{z}) = p(\ybf|\vintbf, \dbf, \bm{z}) \nabla_{\dbf} \log p(\ybf|\vintbf, \dbf, \bm{z}).
\end{equation}
Following the explanations in the main text, we can then compute the gradient of the expectation of $f(\NN)$ as follows,
\begin{align}
\nabla_{\dbf} & \Ej \left[ f(\NN) \right] \\
&= \nabla_{\dbf} \mathbb{E}_{p(\ybf|\vintbf, \dbf, \bm{z}) p(\bm{z}|\vintbf, \dbf) p(\vintbf)} \left[f(\NN)\right] \\
&= \nabla_{\dbf} \mathbb{E}_{p(\ybf|\vintbf, \dbf, \mathbf{h}(\bm{\epsilon}; \vintbf, \dbf)) p(\bm{\epsilon}) p(\bm{\vint}) } \left[f(\NN)\right] \\
&= \int \mathbb{E}_{p(\bm{\epsilon}) p(\vintbf)} \left[\nabla_{\dbf}\{p(\ybf|\vintbf, \dbf, \mathbf{h}(\bm{\epsilon}; \vintbf, \dbf))\} f(\NN)\right] \mathrm{d}\ybf \\
&= \mathbb{E}_{p(\ybf|\vintbf, \dbf, \mathbf{h}(\bm{\epsilon}; \vintbf, \dbf)) p(\bm{\epsilon}) p(\vintbf)} \left[ \nabla_{\dbf}\{\log p(\ybf|\vintbf, \dbf, \mathbf{h}(\bm{\epsilon}; \vintbf, \dbf))\}f(\NN)\right] \label{eq:f_log}
\end{align}
The last equation above can then simply be approximated using a Monte-Carlo sample average. Note that when computing the log derivative $\nabla_{\dbf}\{\log p(\ybf|\vintbf, \dbf, \mathbf{h}(\bm{\epsilon}; \vintbf, \dbf))\}$ we generally require the gradients of the sampling path $\mathbf{h}(\bm{\epsilon}; \vintbf, \dbf)$ as well.
Using Equation 3.9 in the main text, we can derive a similar equation for the derivative of the expectation of $g(\NN)$ over the marginal distribution,
\begin{align}
\nabla_{\dbf} & \Em \left[ g(\NN) \right] \\
&= \nabla_{\dbf} \mathbb{E}_{p(\ybf|\widetilde{\vintbf}, \dbf) p(\widetilde{\vintbf}) p(\vintbf)} \left[g(\NN)\right] \\
&= \nabla_{\dbf} \mathbb{E}_{p(\ybf|\widetilde{\vintbf}, \dbf, \bm{z}) p(\bm{z}|\widetilde{\vintbf}, \dbf) p(\widetilde{\vintbf}) p(\vintbf)} \left[g(\NN)\right] \\
&= \nabla_{\dbf} \mathbb{E}_{p(\ybf|\widetilde{\vintbf}, \dbf, \mathbf{h}(\bm{\epsilon}; \widetilde{\vintbf}, \dbf)) p(\bm{\epsilon}) p(\widetilde{\vintbf}) p(\vintbf) } \left[g(\NN)\right] \\
&= \int \mathbb{E}_{p(\bm{\epsilon}) p(\widetilde{\vintbf}) p(\vintbf)} \left[\nabla_{\dbf}\{p(\ybf|\widetilde{\vintbf}, \dbf, \mathbf{h}(\bm{\epsilon}; \widetilde{\vintbf}, \dbf))\} g(\NN)\right] \mathrm{d}\ybf \\
&= \mathbb{E}_{p(\ybf|\widetilde{\vintbf}, \dbf, \mathbf{h}(\bm{\epsilon}; \widetilde{\vintbf}, \dbf)) p(\bm{\epsilon}) p(\widetilde{\vintbf}) p(\vintbf)} \left[ \nabla_{\dbf}\{\log p(\ybf|\widetilde{\vintbf}, \dbf, \mathbf{h}(\bm{\epsilon}; \widetilde{\vintbf}, \dbf))\} g(\NN)\right], \label{eq:g_log}
\end{align}
where the last equation above can conveniently be approximated using a Monte-Carlo sample average. Equations~\ref{eq:f_log} and~\ref{eq:g_log} do not require differentiating $f$ or $g$ with respect to the neural network output. This allows us to quickly write down gradients of the prominent lower bounds presented in the main text (NWJ, InfoNCE and JSD) when the observation model is known analytically in closed form. This yields Table~\ref{tab:known_grad} in a similar manner to Table 2 in the main text.
\begin{table}[!t]
\begin{minipage}{0.9\textwidth}
\centering
\caption{Gradients of several lower bounds with respect to experimental designs, when the observation model is analytically tractable. We define $T = T_{\bm{\psi}}(\vintbf, \ybf)$ and $T_{ij} = T_{\bm{\psi}}(\bm{\vint}_j, \ybf_i)$, as well as the shorthands $q_{\ybf}=p(\ybf|\vintbf, \dbf, \mathbf{h}(\bm{\epsilon}; \vintbf, \dbf))$, $\widetilde{q}_{\ybf}=p(\ybf|\widetilde{\vintbf}, \dbf, \mathbf{h}(\bm{\epsilon}; \widetilde{\vintbf}, \dbf))$, $P=q_{\ybf} p(\bm{\epsilon}) p(\vintbf)$ and $Q=\widetilde{q}_{\ybf} p(\bm{\epsilon}) p(\widetilde{\vintbf}) p(\vintbf)$. Here, $\text{sp}(T)$ is the softplus function.}
\vspace{3mm}
\begin{tabular}{lll} \toprule
Lower Bound & Gradients with respect to designs \\ \midrule
NWJ         & $\mathbb{E}_{P} \left[ \nabla_{\dbf}\{\log q_{\ybf}\}T\right] - e^{-1} \mathbb{E}_{Q} \left[ \nabla_{\dbf}\{\log \widetilde{q}_{\ybf}\} e^T \right]$ \\
InfoNCE     & $\mathbb{E}_{P^K} \left[ \nabla_{\dbf}\{\log q_{\ybf}\} \frac{1}{K} \sum_{i=1}^K \log{\frac{\exp(T_{ii})}{\frac{1}{K} \sum_{j=1}^K \exp(T_{ij})}} \right]$ \\ 
JSD         & $\mathbb{E}_{P} \left[-\nabla_{\dbf}\{\log q_{\ybf}\} \text{sp}(-T) \right] - \mathbb{E}_{Q} \left[\nabla_{\dbf}\{\log \widetilde{q}_{\ybf}\} \text{sp}(T) \right]$ \\ \bottomrule
\end{tabular}
\label{tab:known_grad}
\end{minipage}
\end{table}
%


\section{Toy experiments}

\subsection{Reference computations}

In order to compute reference values of the mutual information (MI) and posteriors, we rely on an approximation of the likelihood $p(\ybf | \thetab_m, m, \dbf)$ that is based on kernel density estimation (KDE), where $m$ is a model indicator and $\thetab_m$ are the corresponding model parameters.

As described in Equation 4.1 of the main text, the sampling paths of all toy models have additive Gaussian noise $\epsilon \sim \mathcal{N}(\epsilon; 0,1)$ and additive Gamma noise $\nu \sim \Gamma(\nu; 2, 2)$. The overall noise distribution $p_{\text{noise}}$ of $\epsilon + \nu$ is given by a convolution of the individual densities and could be computed via numerical integration. We here opt for an alternative approach, namely to compute the overall noise distribution by a KDE of $50{,}000$ samples of $\epsilon$ and $\nu$. Once we have fitted a KDE to noise samples and obtained an estimate $\widehat{p}_{\text{noise}}$, we can re-arrange the sampling paths, which then allows us to compute estimates of the likelihood. For the set of toy models we here make the assumption that performing experiments with certain designs does not change the data-generating distribution, thereby allowing us to write
\begin{equation}
    \widehat{p}(\ybf|\thetab_m, m, \dbf) = \prod_{j=1}^D \widehat{p}(y_j|\thetab_m, m, d_j),
\end{equation}
where $y_j$ and $d_j$ are the elements of $\ybf$ and $\dbf$, respectively. We can then focus on estimating the likelihood for each dimension of $\dbf$ separately and then multiplying them. Using the estimated noise distribution, the individual likelihood estimates $\widehat{p}(y_j|\thetab_m, m, d_j)$ for the linear ($m=1$), logarithmic ($m=2$) and square-root ($m=3$) toy models are,
\begin{equation} \label{eq:toy_models}
\widehat{p}(y_j|\thetab_m, m, d_j) = 
\begin{cases}
\widehat{p}_{\text{noise}}(y_j - (\theta_{1,0} + \theta_{1,1} d_j)) & \text{if } m = 1, \\
\widehat{p}_{\text{noise}}(y_j - (\theta_{2,0} + \theta_{2,1} \log (|d_j|))) & \text{if } m = 2, \\
\widehat{p}_{\text{noise}}(y_j - (\theta_{3,0} + \theta_{3,1} \sqrt{|d_j|})) & \text{if } m = 3.
\end{cases}
\end{equation}
Below we explain how to use this likelihood approximation to compute reference MI and posteriors for each of the scientific tasks considered in our paper. 
Note that, in the main text, the equations of mutual information are generally given in terms of the posterior to prior ratio. We here compute reference MI values by estimating the ratio of likelihood to marginal, which is equivalent to the posterior to prior ratio by Bayes' rule.

\paragraph{Parameter Estimation} Here we only consider the linear model, where $m=1$, with the aim of solving the task of estimating the model parameters $\thetab_1$, i.e.~the slope and offset of a straight line. The analytic MI is given by Equation 2.5 in the main text and we here approximate it with a nested Monte-Carlo (MC) sample average and the approximate likelihoods in Equation~\ref{eq:toy_models}, i.e.
\begin{equation}
    \widehat{I}(\thetab;\ybf|\dbf) \approx \frac{1}{N} \sum_{i=1}^N \log \frac{\widehat{p}(\ybf^{(i)}|\dbf, \thetab_{1}^{(i)}, m=1)}{\frac{1}{K} \sum_{k=1}^K \widehat{p}(\ybf^{(i)}|\dbf, \thetab_{1}^{(k)}, m=1)},
\end{equation}
where $\ybf^{(i)} \sim p(\ybf | \dbf, \thetab_{1}^{(i)}, m=1)$, $\thetab_{1}^{(i)} \sim p(\thetab_{1})$ and $\thetab_{1}^{(k)} \sim p(\thetab_{1})$. The Gaussian prior distributions for each model are provided in the main text. To compute the above estimate, we use $N=2{,}000$ and $K=500$. We can also use the approximate likelihood and Bayes' rule to compute a reference posterior,
\begin{equation} \label{eq:post_lin}
    \widehat{p}(\thetab_{1}|\dbf, \ybf) = p(\thetab_{1}) \frac{\widehat{p}(\ybf|\dbf, \thetab_{1}, m=1)}{\widehat{p}(\ybf|\dbf)},
\end{equation}
where the marginal $\widehat{p}(\ybf|\dbf) = \int \widehat{p}(\ybf|\dbf, \thetab_{1}, m=1) p(\thetab_{1}) \mathrm{d}\thetab_{1}$ can be computed with a MC sample average or via numerical integration. We here use numerical integration by means of Simpson's rule because of the low dimensions of $\thetab_{1}$.

\paragraph{Model Discrimination} We can approximate the analytic MI in Equation 2.6 of the main text in a similar manner. We use a nested MC sample average, sum over the model indicators $m$ and approximate the relevant likelihoods using Equation~\ref{eq:toy_models}, yielding
\begin{equation}
    \widehat{I}(m;\ybf|\dbf) \approx \sum_{m=1}^3 p(m) \frac{1}{N} \sum_{i=1}^N \log \frac{\widehat{p}(\ybf^{(i)}|\dbf, m)}{\widehat{p}(\ybf^{(i)}|\dbf)},
\end{equation}
where $\ybf^{(i)} \sim p(\ybf|m, \dbf)$ and we use $N=3{,}000$. We approximate the density $\widehat{p}(\ybf^{(i)}|m,\dbf)$ with a MC sample average as well, i.e.
\begin{equation}
    \widehat{p}(\ybf^{(i)}|\dbf, m) = \frac{1}{K} \sum_{k=1}^K \widehat{p}(\ybf^{(i)}|\thetab_{m}^{(k)}, m,  \dbf), \quad \thetab_{m}^{(k)} \sim p(\thetab_m),
\end{equation}
where we use $K=1000$. The density $\widehat{p}(\ybf^{(i)}| \dbf)$ is computed by summing over $m$,
\begin{equation}
    \widehat{p}(\ybf^{(i)}|\dbf) = \sum_{m=1}^3 p(m) \, \widehat{p}(\ybf^{(i)}|\dbf, m).
\end{equation}
Using the above approximate densities and Bayes' rule we can then compute a reference posterior over the model indicator as well, i.e
\begin{equation}
    \widehat{p}(m|\dbf, \ybf) = p(m) \frac{\widehat{p}(\ybf|\dbf, m)}{\widehat{p}(\ybf|\dbf)}.
\end{equation}

\paragraph{Joint MD/PE} The analytic MI for the joint MD/PE task is shown in Equation 2.7 of the main text. A nested MC sample average of that utility is
\begin{equation}
    \widehat{I}(\thetab_m, m;\ybf|\dbf) \approx \sum_{m=1}^3 p(m) \frac{1}{N} \sum_{i=1}^N \log \frac{\widehat{p}(\ybf^{(i)}|\thetab_m^{(i)}, m, \dbf)}{\widehat{p}(\ybf^{(i)}|\dbf)},
\end{equation}
where $\ybf^{(i)} \sim p(\ybf |\thetab_{m}^{(i)}, m,  \dbf)$ and $\thetab_{m}^{(i)} \sim p(\thetab_{m})$. The marginal density $\widehat{p}(\ybf^{(i)}|\dbf)$ is computed by summing over $m$ and using a MC sample average over $\thetab_m$, i.e.
\begin{equation}
    \widehat{p}(\ybf^{(i)}|\dbf) = \sum_{i=1}^3 p(m) \frac{1}{K} \sum_{k=1}^K \widehat{p}(\ybf^{(i)}|\thetab_{m}^{(k)}, m,  \dbf), \quad \thetab_{m}^{(k)} \sim p(\thetab_m).
\end{equation}
In our computations we use $N=2{,}000$ and $K=500$. As before, we can compute a reference joint posterior by using Bayes' rule:
\begin{equation}
    \widehat{p}(\thetab_m, m|\dbf, \ybf) = p(m) \frac{\widehat{p}(\ybf|\thetab_m, m, \dbf)}{\widehat{p}(\ybf|\dbf)}.
\end{equation}


\paragraph{Improving Future Predictions} We here consider the task of improving future predictions for the linear toy model ($m=1$). In order to  approximate the analytic MI shown in Equation 2.8 of the main text, we need to compute approximations of the posterior predictive distribution $\widehat{p}(y_T | \ybf, \dbf, d_T)$, which is given by
\begin{equation}
\widehat{p}(y_T | \ybf, \dbf, d_T) = \int \widehat{p}(y_T | \thetab_{1}, d_T) \widehat{p}(\thetab_{1} | \ybf, \dbf) \mathrm{d} \thetab_{1},
\end{equation}
where $\widehat{p}(y_T | \thetab_{1}, d_T)$ is computed using Equation~\ref{eq:toy_models} and $\widehat{p}(\thetab_{1} | \ybf, \dbf)$ is computed using Equation~\ref{eq:post_lin}. We then approximate $\widehat{p}(y_T | \ybf, \dbf, d_T)$ by numerically integrating over a grid of $\thetab_{1}$. We found that a $50\times50$ grid of $\thetab_1 \in [-10, 10]^2$ was sufficient. In the same manner, we approximate the prior predictive $\widehat{p}(y_T | d_T) = \int \widehat{p}(y_T | \thetab_{1}, d_T) p(\thetab_{1}) \mathrm{d}\thetab_1$ by numerical integration. Being able to compute the posterior predictive for a certain $\{\dbf, \ybf\}$ then allows us to approximate the analytic MI using a combination of numerical integration and MC sample averages, i.e.
\begin{equation*}
\widehat{I}(y_T|d_T;\ybf|\dbf) \approx \frac{1}{N} \sum_{i=1}^N \int \widehat{p}(y_T | \ybf^{(i)}, \dbf, d_T) \log \frac{\widehat{p}(y_T | \ybf^{(i)}, \dbf, d_T)}{\widehat{p}(y_T | d_T)} \mathrm{d}y_T,
\end{equation*}
where $\ybf^{(i)} \sim p(\ybf|\dbf)$. We used $N=800$ and numerically integrated over $y_T \in [-50, 50]$ using Simpson's rule with a grid of size $100$.

\subsection{Data distributions}

\begin{figure}[!t]
    \centering
    \includegraphics[width=\linewidth]{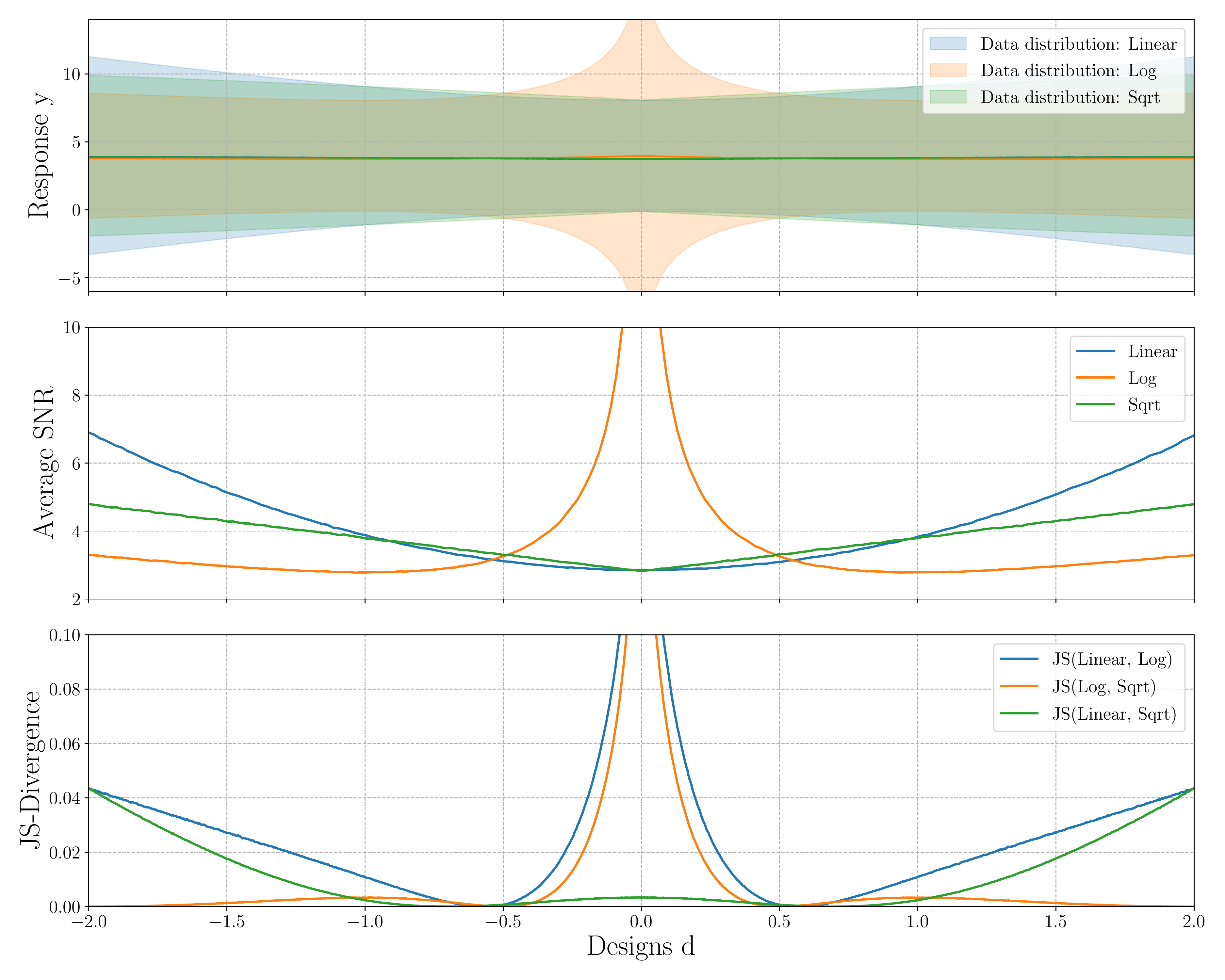}
    \vspace{-0.7cm}
    \caption[]{Toy model data distributions (top), average signal-to-noise ratio (SNR, middle) and Jensen-Shannon (JS) divergences between different data distributions (bottom).}
    \label{fig:toy_data}
\end{figure}

In Figure~\ref{fig:toy_data} we summarise general information about data simulated from each toy model presented in the main text, i.e.~the linear model, the log model and the square-root model. The top row shows the prior predictive distribution for each model, where the solid lines represent the means and the shaded areas indicate one standard deviation from the means. While the linear and square-root model have a similar data distribution, the log model prior predictive is significantly different, rapidly increasing as $d$ approaches $0$. This signifies in which regions the data distributions are most different, under our prior belief about the model parameters $\thetab_m$, i.e.~at the boundaries $d=-2, 2$ and at $d=0$. The bottom plot, which shows the Jensen-Shannon (JS) divergence between the data distributions of different toy models, further emphasises this difference. This is in accordance with the optimal designs found for the toy model MD and MD/PE tasks in the main text.

The middle row of Figure~\ref{fig:toy_data} shows the average signal-to-noise ratio (SNR) for each of the toy models. The SNR, for a single $\thetab_m$, is given by
\begin{equation} \label{eq:snr}
    \text{SNR}(d|\thetab_m) = \left(\frac{\mu(d|\thetab_m)}{\sigma(d|\thetab_m)} \right)^2,
\end{equation}
where $\mu(d|\thetab_m)$ and $\sigma(d|\thetab_m)$ are the mean and standard deviation of the model response $y|\thetab_m, d$, respectively, averaged over signal noise. The average SNR is then $\overline{\text{SNR}}(d) = \mathbb{E}_{p(\thetab)}[\text{SNR}(d|\thetab_m)]$.
While the linear and square-root model have a relatively high average SNR at the boundaries, the log model has an extremely high average SNR at $d=0$. A high SNR is useful for parameter estimation, because the observed data essentially has less \emph{relative} noise, allowing us to better identify which values of $\thetab_m$ might have generated observations. This is again in accordance with results found in the main text. 

\subsection{Architectures and hyper-parameters}

In Table~\ref{tab:toy_hp} we show neural network (NN) architectures, including the number of hidden layers and number of hidden units, as well as the learning rates (L.R.) for the NN parameters $\psib$ and experimental designs $\dbf$. We optimise $\psib$ and $\dbf$ with two separate Adam optimisers. with default parameters from the PyTorch package in Python. Each epoch, we simulate $10{,}000$ new data samples.



%
\begin{table}[!t]
\begin{minipage}{0.9\textwidth}
\centering
\caption{Toy model hyper-parameters for each scientific goal.}
\vspace{3mm} 
\begin{tabular}{lcccc} \toprule
Scientific Goal & NN Layers & NN Units & L.R. for $\psib$ & L.R. for $\dbf$ \\ \midrule
PE              & 2         & 50      & $10^{-4}$        & $10^{-3}$       \\
MD              & 2         & 50       & $10^{-3}$        & $10^{-4}$       \\
MD/PE           & 2         & 70       & $10^{-4}$        & $5\times10^{-4}$       \\
FP              & 2         & 100      & $10^{-3}$        & $10^{-2}$       \\ \bottomrule
\end{tabular}
\label{tab:toy_hp}
\end{minipage}
\end{table}
%


\subsection{Additional results}

\begin{figure}[!t]
    \centering
    \includegraphics[width=\linewidth]{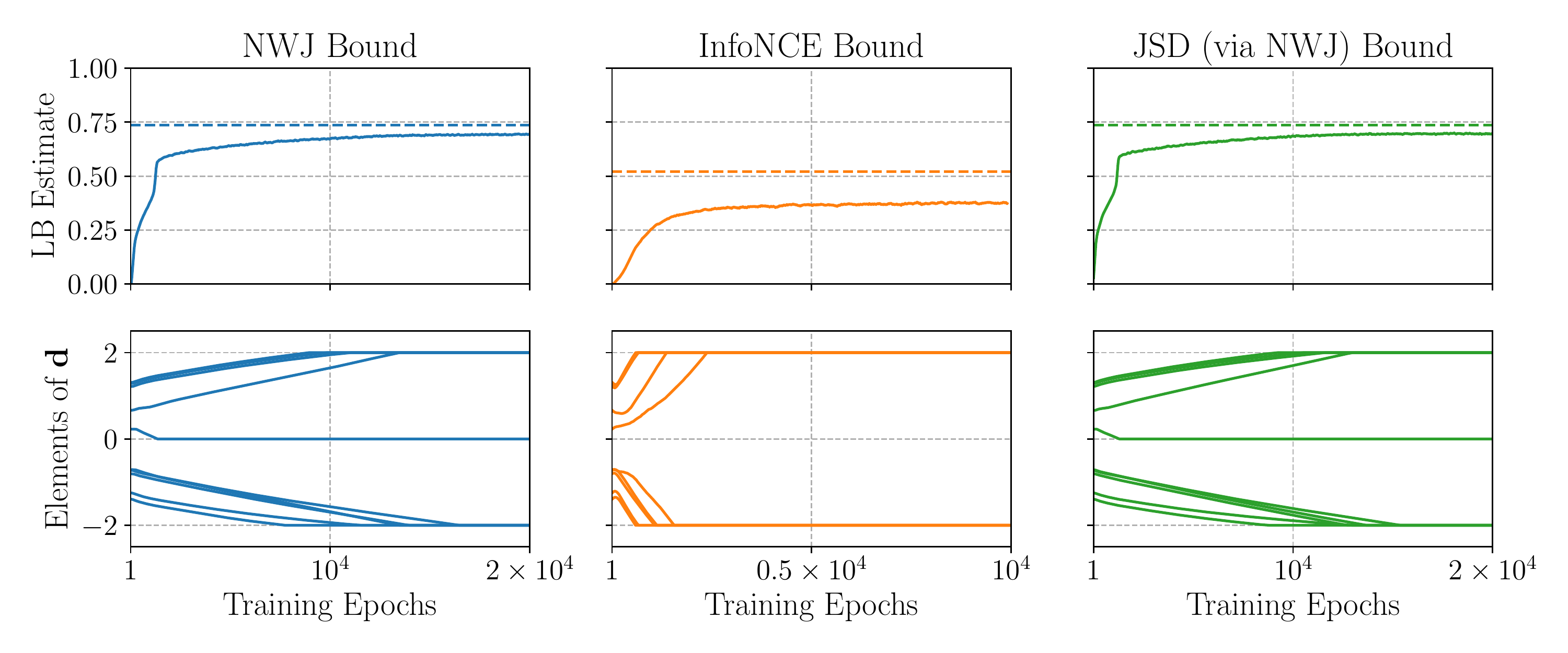}
    \vspace{-0.7cm}
    \caption[]{MD training curves when InfoNCE yields different locally optimal design. The top row shows lower bounds as a function of training epochs, with the dotted line being the reference MI at the final optimal design of the respective lower bound. The bottom row shows the corresponding design elements as the design vector is being updated.}
    \label{fig:md_rerun}
\end{figure}

Here we show additional results for the toy model experiments that were mentioned in the main text. Figure~\ref{fig:md_rerun} shows a setting where the goal is MD and the InfoNCE lower bound found different optimal experimental designs than the NWJ and JSD lower bound. This is slightly surprising, as the neural networks for each lower bound used the same architecture and the same initialisation. Because different optimal designs were found, the final MI estimate for the InfoNCE lower bound is lower than that of the NWJ and JSD lower bounds. Importantly, this implies that the choice of lower bound does matter, as different optimal designs may be found. In our work we do not, however, aim to explore which lower bound is best, but rather show how different lower bounds may be used in our proposed framework.

\begin{figure}[!t]
    \centering
    \includegraphics[width=\linewidth]{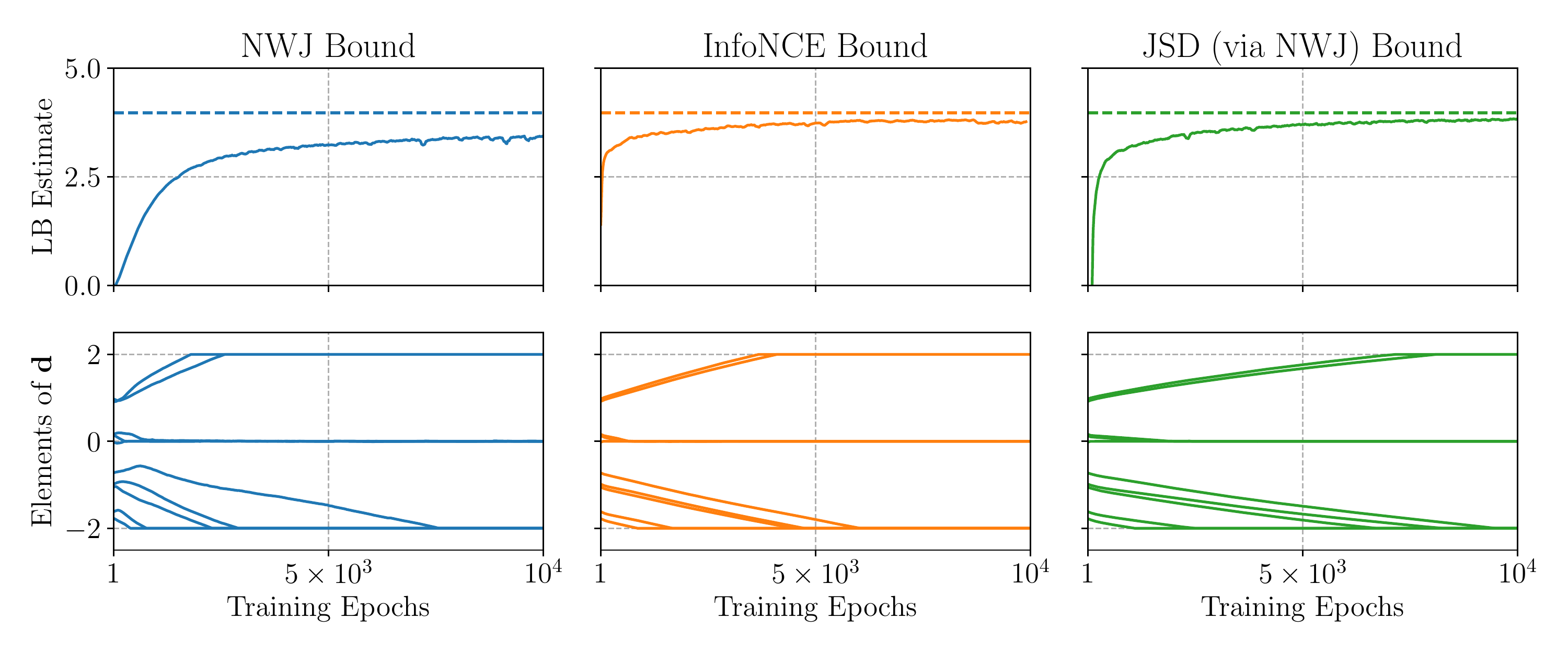}
    \vspace{-0.7cm}
    \caption[]{MD/PE results for the set of toy models. Shown are training curves for the NWJ, InfoNCE and JSD (via NWJ) lower bounds. The top row shows lower bound estimates as a function of training epochs, with the dotted line being the reference MI value, and the bottom row shows the elements of the design vector as it is being updated.}
    \label{fig:mdpe_training}
\end{figure}

\begin{figure}[!t]
    \centering
    \includegraphics[width=\linewidth]{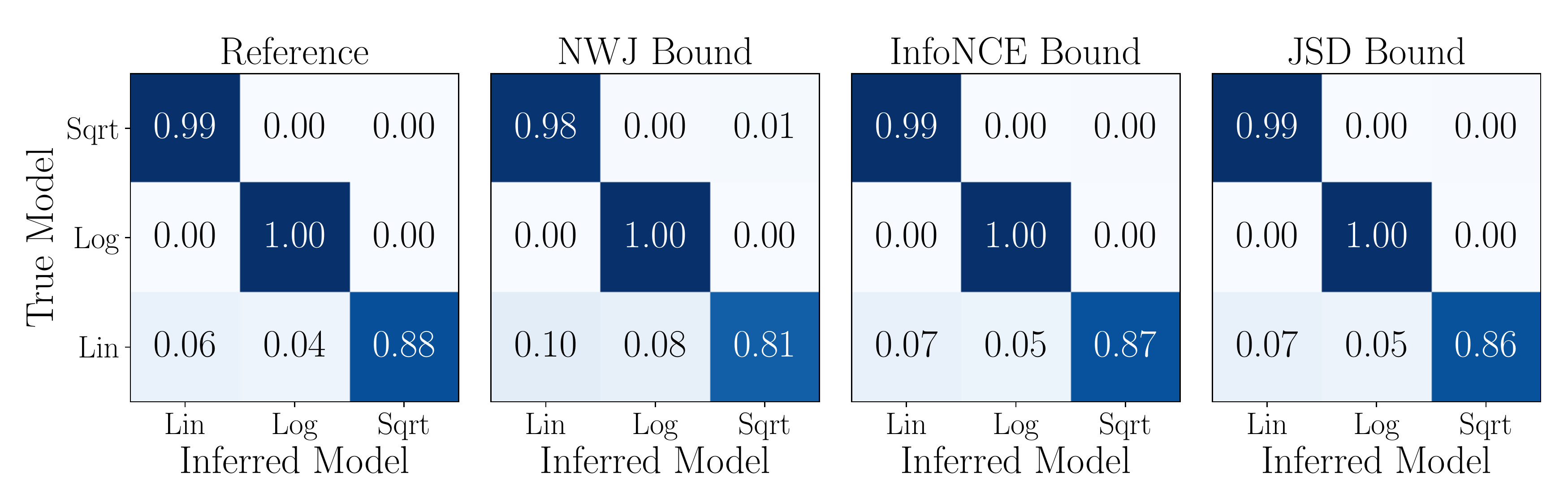}
    \vspace{-0.7cm}
    \caption[]{MD/PE posterior results of the model indicator for the set of toy models. Shown are average posterior probabilities for different ground truths (one per row) for the NWJ, InfoNCE and JSD lower bounds.}
    \label{fig:mdpe_posterior_indic}
\end{figure}

\begin{figure}[!t]
    \centering
    \includegraphics[width=\linewidth]{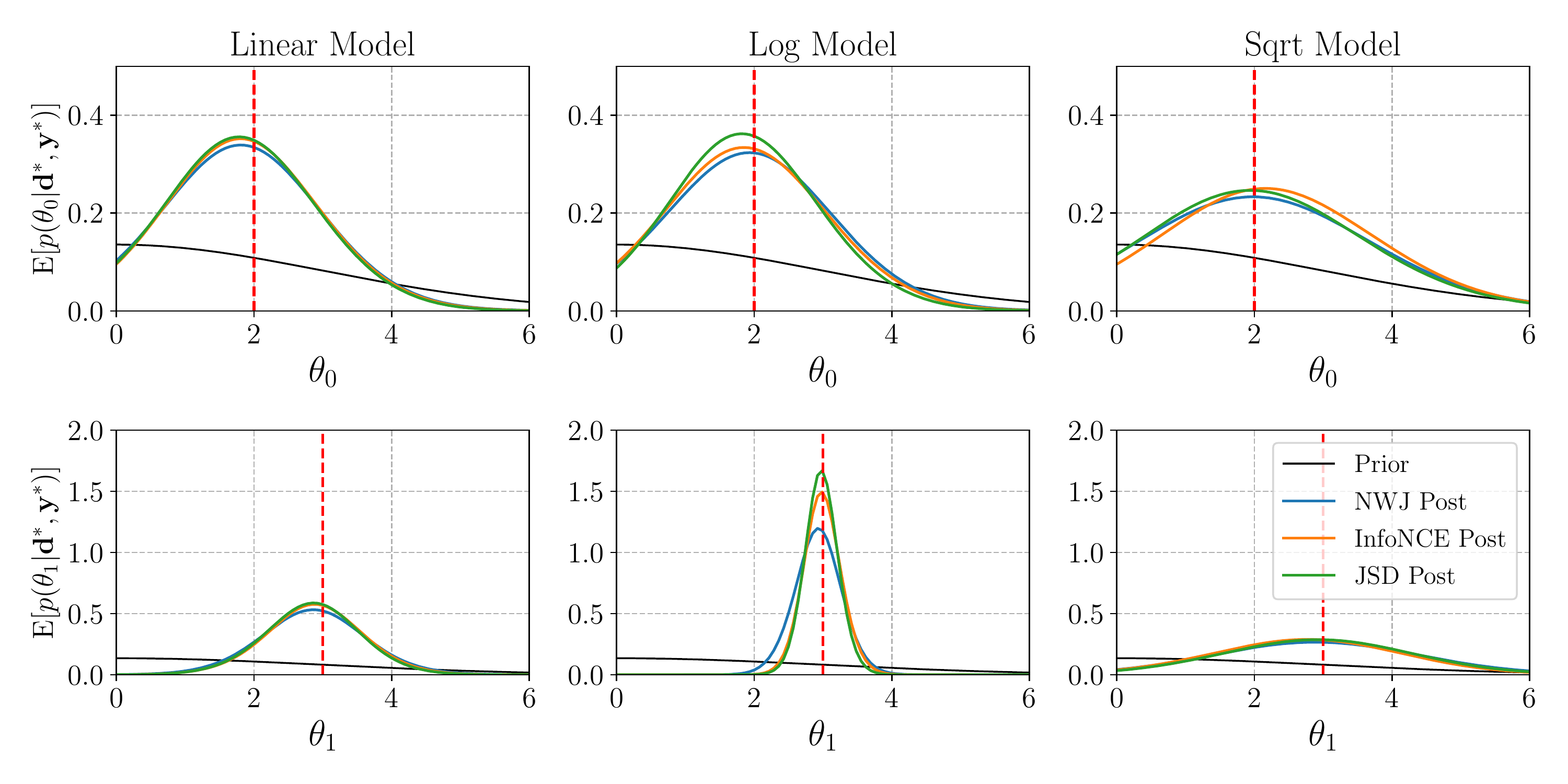}
    \vspace{-0.7cm}
    \caption[]{MD/PE posterior results of the model parameters for all toy models. Shown are average marginal posteriors for all lower bounds, including the prior and an average reference posterior.}
    \label{fig:mdpe_posterior_param}
\end{figure}

\begin{figure}[!t]
    \centering
    \includegraphics[width=\linewidth]{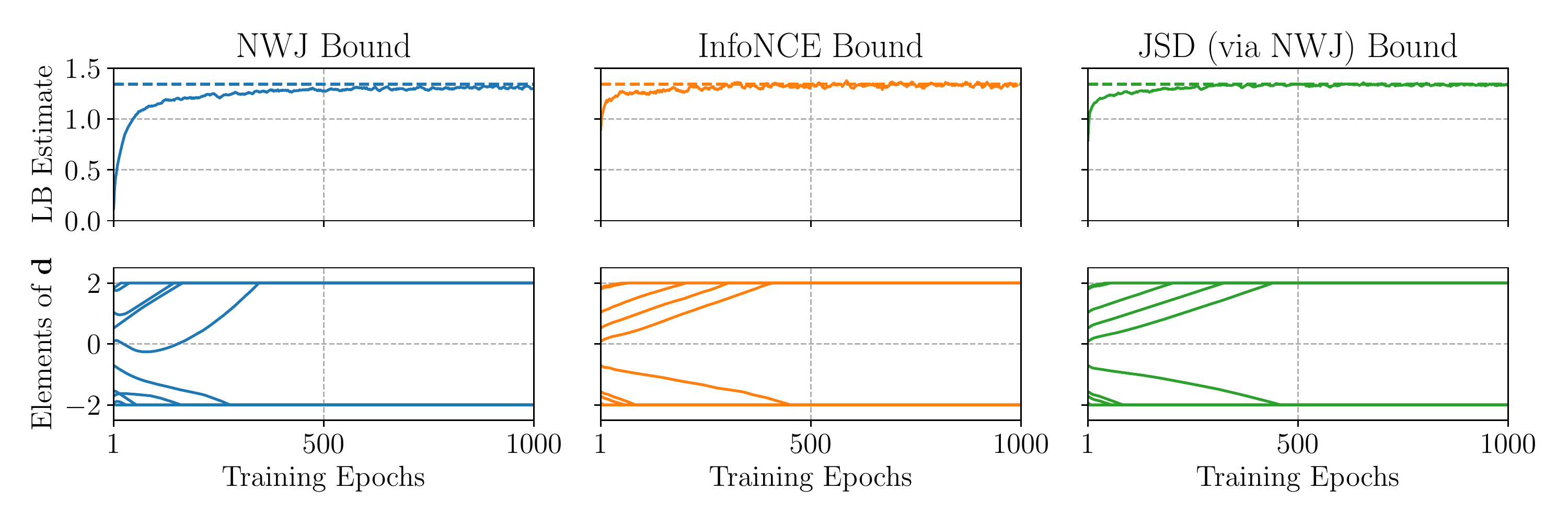}
    \vspace{-0.7cm}
    \caption[]{FP results for the linear toy model. Shown are training curves for the NWJ, InfoNCE and JSD (via NWJ) lower bounds. The top row shows lower bound estimates as a function of training epochs, with the dotted line being the reference MI value, and the bottom row shows the elements of the design vector as it is being updated.}
    \label{fig:fp_training}
\end{figure}

In Figure~\ref{fig:mdpe_training} we show the training curves of each lower bound (top row) for the joint MD/PE task, as well as the experimental designs as they are being updated (bottom row). All lower bounds yield the same optimal design and converge to a final MI close to a reference MI (with the NWJ bound being further away). Using the trained neural networks for each lower bound we can compute posterior distributions, as described in the main text. Figure~\ref{fig:mdpe_posterior_indic} shows marginal posterior distributions over the model indicator $m$ and Figure~\ref{fig:mdpe_posterior_param} shows marginal posterior distributions over model parameters $\thetab_m$. Note that we show average posterior distributions over $5{,}000$ `real-world' observations $\ybf^\ast |\dbf^\ast, \thetab_{m,\text{true}}$, where $\thetab_{m,\text{true}} = (2, 3)$ for all models. Finally, for the FP task, we show the training curves of each lower bound for the FP task (top row), as well as corresponding design curves (bottom row), in Figure~\ref{fig:fp_training}.

\clearpage

\section{Epidemiology experiments}

\subsection{Discrete state continuous time Markov chain formulation}

We start by defining the SIR and SEIR models in terms of continuous time Markov chains (CTMC), where time is continuous and the state populations are discrete.

Recall that the SIR model~\citep[e.g.][]{Allen2008} is governed by the state changes $S_1(t) \rightarrow I_1(t) \rightarrow R_1(t)$, where $S_1(t)$ are the number of susceptible individuals, $I_1(t)$ are the number of infectious individuals and $R_1(t)$ are the individuals that have recovered and can no longer be infected. The dynamics of these state changes are determined by the model parameters $\thetab_1 = (\beta, \gamma)$, where $\beta$ is the infection rate and $\gamma$ is the recovery rate. As done in the main text, let us define a population vector $\mathbf{X}_1(t) = (S_1(t), I_1(t))^\top$.\footnote{We can ignore the population of recovered individuals here because we assume that the total population $N$ stays constant.} The state changes $\Delta \mathbf{X}_1(t)$ of the SIR model within a small time frame $\Delta t$ are then affected by two events: \emph{infection}, where $\Delta \mathbf{X}_1(\Delta t) = (-1, +1)^\top$, and \emph{recovery}, where $\Delta \mathbf{X}_1(\Delta t) = (0, -1)^\top$. Denoting the $i$-th event by $(\Delta \mathbf{X}_1(t))^i$, Table~\ref{tab:sir_changes} summarises these state changes and their corresponding transition probabilities\footnote{$\smallO(x)$ in Tables~\ref{tab:sir_changes} and~\ref{tab:seir_changes} denotes a function that goes faster to zero than $x$, i.e. $\lim_{x\to 0} \smallO(x)/x = 0$.}, which define the behaviour of the CTMC SIR model.
\begin{table}[!t]
\begin{minipage}{1\textwidth}
\centering
\caption{CTMC SIR model state changes and their probabilities.}
\vspace{3mm} 
\begin{tabular}{lll} \toprule
Event & State Change & Probability \\ \midrule
Infection & $(\Delta \mathbf{X}_1(t))^1 = (-1, +1)^\top$ & $p_1 = \beta \frac{S_1(t)I_1(t)}{N} \Delta t + \smallO(\Delta t) $\\
Recovery & $(\Delta \mathbf{X}_1(t))^2 = (\,\,\,\,\,0, -1)^\top$ & $p_2 = \gamma I_1(t) \Delta t + \smallO(\Delta t) $\\\bottomrule
\end{tabular}
\label{tab:sir_changes}
\end{minipage}
\end{table}

The SEIR model~\citep[][]{Lekone2006} introduces an exposed state $E_2(t)$, where individuals are infected but not yet infectious, and is governed by the state changes $S_2(t) \rightarrow E_2(t)  \rightarrow  I_2(t) \rightarrow R_2(t)$. The dynamics of these changes are determined by the model parameters $\thetab_2 = (\beta, \gamma, \sigma)$, where $\beta$ and $\gamma$ are as before and $\sigma^{-1}$ is the average incubation period. We again define a population vector $\mathbf{X}_2(t) = (S_2(t), E_2(t), I_2(t))^\top$. The state changes $\Delta \mathbf{X}_2(t)$ of the SEIR model are affected by three events: \emph{infection}, \emph{becoming infectious} and \emph{recovery}. These state changes, denoted by $(\Delta \mathbf{X}_2(t))^i$, and their corresponding probabilities, which together define the CTMC SEIR model, are summarised in Table~\ref{tab:seir_changes}.
\begin{table}[!t]
\begin{minipage}{1\textwidth}
\centering
\caption{CTMC SEIR model state changes and their probabilities.}
\vspace{3mm} 
\begin{tabular}{lll} \toprule
Event & State Change & Probability \\ \midrule
Infection & $(\Delta \mathbf{X}_2(t))^1 = (-1, +1, \,\,\,\,\,0)^\top$ & $p_1 = \beta \frac{S_2(t)I_2(t)}{N} \Delta t + \smallO(\Delta t)$ \\
Becoming infectious & $(\Delta \mathbf{X}_2(t))^2 = (\,\,\,\,\,0, -1, +1)^\top$ & $p_2 = \sigma E_2(t) \Delta t + \smallO(\Delta t)$ \\
Recovery & $(\Delta \mathbf{X}_2(t))^3 = (\,\,\,\,\,0,\,\,\,\,\,0, -1)^\top$ & $p_3 = \gamma I_2(t) \Delta t + \smallO(\Delta t)$ \\\bottomrule
\end{tabular}
\label{tab:seir_changes}
\end{minipage}
\end{table}

\subsection{Diffusion Approximations}

Based on the CTMC formulation of the SIR and SEIR models in Table~\ref{tab:sir_changes} and Table~\ref{tab:seir_changes}, we can derive continuous-state diffusion approximations of these epidemiological processes using the method of~\citet{E_Allen2008}, which results in systems of stochastic differential equations (SDEs).

In the main text we 
wrote down the system of Itô SDEs as
\begin{equation} \label{eq:sde}
\mathrm{d}\mathbf{X}_m(t) = \mathbf{f}_m(\mathbf{X}_m(t)) \mathrm{d}t + \mathbf{G}_m(\mathbf{X}_m(t))\mathrm{d}\mathbf{W}(t),
\end{equation}
where $m$ is a model indicator, $\mathbf{f}_m$ the drift vector, $\mathbf{G}_m$ the diffusion matrix, and $\mathbf{W}(t)$ is a vector of independent Wiener processes. The drift vector equals the infinitesimal mean $\lim_{\mathrm{d}t \to 0}\mathbb{E}(\mathrm{d}\mathbf{X}_m(t))/\mathrm{d}t$ of the diffusion, and $\mathbf{G}_m$ defines the infinitesimal variance $\lim_{\mathrm{d}t \to 0}\mathbb{V}\text{ar}(\mathrm{d}\mathbf{X}_m(t))/\mathrm{d}t = \mathbf{G}_m\mathbf{G}_m^\top $. They can be chosen such that the diffusion approximation matches the infinitesimal mean and variance of the discrete state continuous Markov chain models. ~\citet{E_Allen2008} provide a systematic way for doing this, which we shall follow below.

Let us first consider the CTMC SIR model summarised in Table~\ref{tab:sir_changes}. The drift vector is chosen to match the expected change in $\mathbf{X}_1(t)$ of the CTMC formulation, i.e.
\begin{align}
\mathbf{f_1}(\mathbf{X}_1(t)) 
&= \lim_{\Delta t \to 0}\frac{1}{\Delta t} \mathbb{E}\left[ \Delta \mathbf{X}_1(t) \right] \\
&= \lim_{\Delta t \to 0}\frac{1}{\Delta t} \sum_{i=1}^2 p_i (\Delta \mathbf{X}_1(t))^i \\
&= \beta \frac{S_1(t)I_1(t)}{N} \begin{pmatrix} -1 \\ 1 \end{pmatrix} + \gamma I_1(t) \begin{pmatrix}0 \\ -1 \end{pmatrix} \\
&= \begin{pmatrix}
        -\beta \frac{S_1(t)I_1(t)}{N} \\[1em]
        \beta \frac{S_1(t)I_1(t)}{N} - \gamma I_1(t) \\[1em]
    \end{pmatrix}.
\end{align}
In order to derive the diffusion matrix $\mathbf{G}_1$, let us first define a matrix $\bm{\Lambda}_1$ whose rows correspond to the state changes $(\Delta \mathbf{X}_1(t))^i$ in Table~\ref{tab:sir_changes}, i.e.
\begin{equation}
    \bm{\Lambda}_1 =
        \begin{pmatrix}
            -1 & +1 & \\
             0 & -1 & \\
        \end{pmatrix}.
\end{equation}
~\citet{E_Allen2008} show that the elements $G_{1, ij}$ of the diffusion matrix $\mathbf{G}_1$ are then given by
\begin{equation} \label{eq:lambda}
    G_{1, ij} = \Lambda_{ji} \, p_j^{1/2},
\end{equation}
allowing us to derive the full diffusion matrix using Table~\ref{tab:sir_changes} and Equation~\ref{eq:lambda}, i.e.
\begin{align}
\mathbf{G}_1(\mathbf{X}_1(t)) 
&= \begin{pmatrix}
        \Lambda_{11} \, p_1^{1/2} & \Lambda_{21} \, p_2^{1/2} \\
        \Lambda_{12} \, p_1^{1/2} & \Lambda_{22} \, p_2^{1/2} \\
        \end{pmatrix} \\
&=  \begin{pmatrix}
        - \sqrt{\beta \frac{S_1(t)I_1(t)}{N}} & 0 \\[1em]
        \sqrt{\beta \frac{S_1(t)I_1(t)}{N}}   & - \sqrt{\gamma I_1(t)} \\[1em]
    \end{pmatrix}.
\end{align}
We can check that this diffusion matrix matches the infinitesimal variance of the CTMC SIR model, i.e.~
\begin{align}
\lim_{\Delta t \to 0}\frac{\mathbb{V}\text{ar}(\Delta\mathbf{X}_1(t))}{\Delta t}
&= \lim_{\Delta t \to 0} \frac{1}{\Delta t}\sum_{i=1}^2 p_i (\Delta \mathbf{X}_1(t))^i (\Delta \mathbf{X}_1(t))^{i, \top} - \lim_{\Delta t \to 0}\frac{1}{\Delta t} \mathbb{E}\left[ \Delta \mathbf{X}_1(t) \right]\mathbb{E}\left[ \Delta \mathbf{X}_1(t) \right]^\top \nonumber \\
&= \lim_{\Delta t \to 0} \frac{1}{\Delta t}\sum_{i=1}^2 p_i (\Delta \mathbf{X}_1(t))^i (\Delta \mathbf{X}_1(t))^{i, \top} - \lim_{\Delta t \to 0} \frac{\smallO(\Delta t)}{\Delta t} \\
&= \beta \frac{S_1(t)I_1(t)}{N} \begin{pmatrix} -1 \\ 1 \end{pmatrix} \begin{pmatrix} -1 & 1 \end{pmatrix} + \gamma I_1(t) \begin{pmatrix} 0 \\ -1 \end{pmatrix} \begin{pmatrix} 0 & -1 \end{pmatrix}\\
&= \begin{pmatrix}
         \beta \frac{S_1(t)I_1(t)}{N} & - \beta \frac{S_1(t)I_1(t)}{N} \\[1em]
         - \beta \frac{S_1(t)I_1(t)}{N} & \beta \frac{S_1(t)I_1(t)}{N} + \gamma I_1(t) \\[1em]
     \end{pmatrix},
\end{align}
which exactly corresponds to $\mathbf{G}_1 \mathbf{G}_1^{\top}$, as required. We note that the diffusion matrix $\mathbf{G}_1$ is not uniquely defined, as we only wish to match the variance. For instance, we could multiply $\mathbf{G}_1$ by any orthogonal matrix and the product $\mathbf{G}_1\mathbf{G}_1^{\top}$ remains unchanged.

We now follow the same procedure for the SEIR model. To do so, let us consider the state changes and corresponding probabilities summarised in Table~\ref{tab:seir_changes}.
%
As done for the SIR model, given these state changes and probabilities, we can derive the drift vector by computing the expected change in $\mathbf{X}_2(t)$,
\begin{align}
\mathbf{f_2}(\mathbf{X}_2(t)) 
&= \lim_{\Delta t \to 0} \frac{1}{\Delta t} \mathbb{E}\left[ \Delta \mathbf{X}_2(t) \right] \\
&= \lim_{\Delta t \to 0} \frac{1}{\Delta t} \sum_{i=1}^3 p_i (\Delta \mathbf{X}_2(t))^i \\
&= \beta \frac{S_2(t)I_2(t)}{N} \begin{pmatrix} -1 \\ 1 \\ 0 \end{pmatrix} + \sigma E_2(t) \begin{pmatrix} 0 \\ -1 \\ 1 \end{pmatrix} +
\gamma I_2(t) \begin{pmatrix} 0 \\ 0 \\ -1 \end{pmatrix} \\
&= \begin{pmatrix}
        -\beta \frac{S_2(t)I_2(t)}{N} \\[1em]
        \beta \frac{S_2(t)I_2(t)}{N} - \sigma E_2(t) \\[1em]
        \sigma E_2(t) - \gamma I_2(t) \\[1em]
    \end{pmatrix}.
\end{align}

\clearpage

In order to derive the diffusion matrix $\mathbf{G}_2$, we define a matrix $\bm{\Lambda}_2$ whose rows correspond to the state changes in Table~\ref{tab:seir_changes}, i.e.
\begin{equation}
    \bm{\Lambda}_2 =
        \begin{pmatrix}
            -1 & +1 & 0 \\
             0 & -1 & +1 \\
             0 &  0 & -1 \\
        \end{pmatrix}.
\end{equation}
Substituting $\bm{\Lambda}_2$ in Equation~\ref{eq:lambda} and using Table~\ref{tab:seir_changes} then allows us to derive $\mathbf{G}_2$,
%
\begin{align}
\mathbf{G}_2(\mathbf{X}_2(t)) 
&= \begin{pmatrix}
        \Lambda_{11} \, p_1^{1/2} & \Lambda_{21} \, p_2^{1/2} & \Lambda_{31} \, p_3^{1/2} \\
        \Lambda_{12} \, p_1^{1/2} & \Lambda_{22} \, p_2^{1/2} & \Lambda_{32} \, p_3^{1/2} \\
        \Lambda_{13} \, p_1^{1/2} & \Lambda_{23} \, p_2^{1/2} & \Lambda_{33} \, p_3^{1/2} \\
        \end{pmatrix} \\
&= \begin{pmatrix}
        - \sqrt{\beta \frac{S_2(t)I_2(t)}{N}} & 0 & 0 \\
        \sqrt{\beta \frac{S_2(t)I_2(t)}{N}} & - \sqrt{\sigma E_2(t)} & 0 \\
        0 & \sqrt{\sigma E_2(t)} & - \sqrt{\gamma I_2(t)} \\
        \end{pmatrix}.
\end{align}
We can again check that $G_2$ matches the infitesimal variance of the CTMC SEIR model,
\begin{align}
\lim_{\Delta t \to 0}\frac{\mathbb{V}\text{ar}(\Delta\mathbf{X}_2(t))}{\Delta t}
&= \lim_{\Delta t \to 0} \frac{1}{\Delta t} \sum_{i=1}^3 p_i (\Delta \mathbf{X}_2(t))^i (\Delta \mathbf{X}_2(t))^{i, \top} - \lim_{\Delta t \to 0}\frac{1}{\Delta t} \mathbb{E}\left[ \Delta \mathbf{X}_2(t) \right]\mathbb{E}\left[ \Delta \mathbf{X}_2(t) \right]^\top \nonumber \\
&= \lim_{\Delta t \to 0} \frac{1}{\Delta t} \sum_{i=1}^3 p_i (\Delta \mathbf{X}_2(t))^i (\Delta \mathbf{X}_2(t))^{i, \top} - \lim_{\Delta t \to 0} \frac{\smallO(\Delta t)}{\Delta t} \\
&
\begin{aligned}
= \beta \frac{S_2(t)I_2(t)}{N} &\begin{pmatrix} -1 \\ 1 \\ 0 \end{pmatrix} \begin{pmatrix} -1 & 1 & 0 \end{pmatrix} \\
&+ \sigma E_2(t) \begin{pmatrix} 0 \\ -1 \\ 1 \end{pmatrix} \begin{pmatrix} 0 & -1 & 1\end{pmatrix} \\
&+ \gamma I_2(t) \begin{pmatrix} 0 \\ 0 \\ -1 \end{pmatrix} \begin{pmatrix} 0 & 0 & -1\end{pmatrix} 
\end{aligned} \\
&= \begin{pmatrix}
        \beta \frac{S_2(t)I_2(t)}{N} & -\beta \frac{S_2(t)I_2(t)}{N} & 0 \\[1em]
        -\beta \frac{S_2(t)I_2(t)}{N} & \beta \frac{S_2(t)I_2(t)}{N} + \sigma E_2(t) & -\sigma E_2(t) \\[1em]
        0 & -\sigma E_2(t) & \sigma E_2(t) + \gamma I_2(t) \\
        \end{pmatrix},
\end{align}
which exactly corresponds to $\mathbf{G}_2 \mathbf{G}_2^{\top}$, as required. However, as previously said, the diffusion matrix $\mathbf{G}_2$ is not uniquely defined.

\newpage
\subsection{Architectures and hyper-parameters}

We list the hyper-parameters for the parameter estimation (PE) and model discrimination (MD) task of the SDE-based epidemiology experiments in Table~\ref{tab:sde_pe_hp} and Table~\ref{tab:sde_md_hp}, respectively, for all experimental budgets $D$ that were presented in the main text. Shown are neural network (NN) architectures, including the number of hidden layers and number of hidden units, as well as the learning rates (L.R.) for the NN parameters $\psib$ and experimental designs $\dbf$. We optimise $\psib$ and $\dbf$ with two separate Adam optimisers with default parameters from the PyTorch package in Python. We pre-simulate $20{,}000$ SDE solutions on a fine time grid. During training time we can then access solutions, and their gradients, at a specific time point by simply looking up the solutions corresponding to the nearest point in the time grid. We found that this was generally much faster than simulating/solving the SDEs on the fly, at the cost of higher memory usage.

\begin{table}[!h]
\begin{minipage}{0.9\textwidth}
\centering
\caption{PE hyper-parameters for the SDE-based SIR model, for varying number of measurements $D$ (i.e.~experimental budget).}
\vspace{3mm} 
\begin{tabular}{ccccc} \toprule
Budget $D$ & NN Layers & NN Units & L.R. for $\psib$ & L.R. for $\dbf$ \\ \midrule
1  & 2  & 20 & $10^{-4}$        & $3\times10^{-2}$ \\
2  & 3  & 20 & $10^{-4}$        & $3\times10^{-2}$ \\
3  & 4  & 20 & $10^{-4}$        & $3\times10^{-2}$ \\
5  & 5  & 20 & $3\times10^{-4}$ & $3\times10^{-2}$ \\
10 & 3  & 30 & $10^{-4}$        & $10^{-2}$        \\
\bottomrule
\end{tabular}
\label{tab:sde_pe_hp}
\end{minipage}
\end{table}
\begin{table}[!h]
\begin{minipage}{0.9\textwidth}
\centering
\caption{MD hyper-parameters for the SDE-based SIR and SEIR models, for varying number of measurements $D$ (i.e.~experimental budget).}
\vspace{3mm} 
\begin{tabular}{ccccc} \toprule
Budget $D$ & NN Layers & NN Units & L.R. for $\psib$ & L.R. for $\dbf$ \\ \midrule
1  & 2  & 20 & $10^{-4}$        & $3\times10^{-2}$ \\
2  & 3  & 20 & $10^{-4}$        & $3\times10^{-2}$ \\
3  & 4  & 20 & $10^{-4}$        & $3\times10^{-2}$ \\
5  & 5  & 20 & $3\times10^{-4}$ & $3\times10^{-2}$ \\
10 & 3  & 30 & $10^{-4}$        & $10^{-2}$        \\
20 & 3  & 30 & $10^{-4}$        & $10^{-2}$        \\
\bottomrule
\end{tabular}
\label{tab:sde_md_hp}
\end{minipage}
\end{table}

\pagebreak

\subsection{Data distributions}

\begin{figure}[!h]
    \centering
    \includegraphics[width=\linewidth]{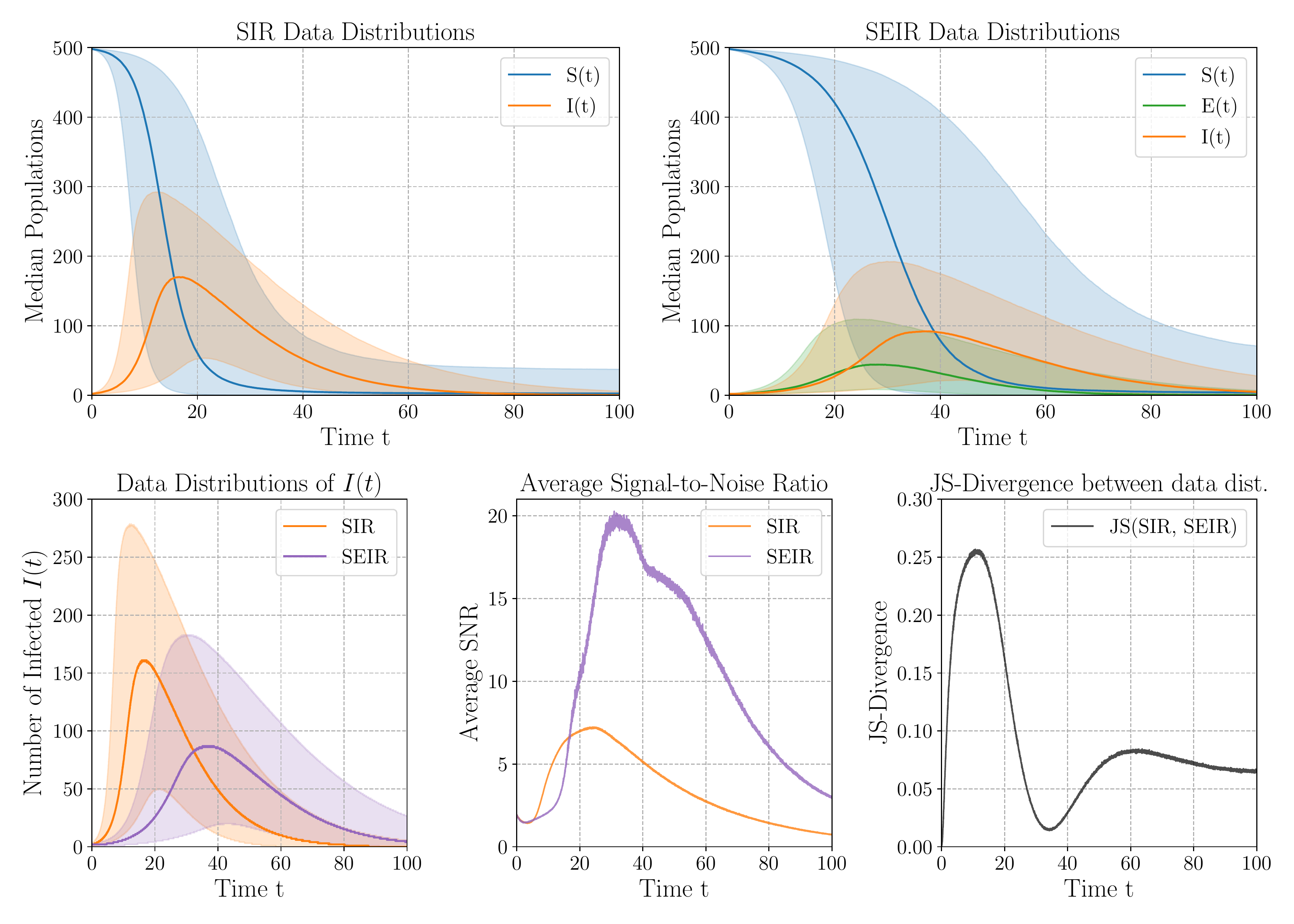}
    \vspace{-0.7cm}
    \caption[]{General summary of data simulated from the SDE-based epidemiology models. The top row shows prior predictive distributions for the SIR model (top left) and the SEIR model (top right). The bottom left figure shows the number of infectious individuals for the SIR and SEIR models, while the bottom centre plot shows the corresponding average signal-to-noise ratios (SNR). The bottom right shows the Jensen-Shannon (JS) divergence between the prior predictive distributions of both models.}
    \label{fig:sde_data}
\end{figure}

In Figure~\ref{fig:sde_data} we provide general information about data simulated from the SIR and SEIR models. The top left plot shows the SIR model prior predictive distributions of the number of susceptible individuals $S(t)$ and the number of infectious individuals $I(t)$ as a function of time $t$. Similarly, the top right plot shows the SEIR model prior predictive distributions of $S(t)$, $I(t)$ and the number of exposed individuals $E(t)$. Recall, however, that we only use the number of infectious individuals as data in our experiments. As such, we compare $I(t)$ from the SIR and from the SEIR model more closely in the bottom left plot. As can be seen, the number of infectious individuals for the SEIR model peak at later times $t$. The bottom centre plot then shows the average signal-to-noise ratio (SNR), computed by means of Equation~\ref{eq:snr}, for the $I(t)$ response of the SIR and SEIR model. The average SNR for the SEIR model is generally much higher than that of the SIR model, also peaking at later measurement times. Lastly, the bottom right plot shows the Jensen-Shannon divergence between the prior predictives of the SIR and SEIR model, i.e.~$\text{JS}(p(I_1(t)|t) \mid\mid p(I_2(t)|t))$, where $I_1(t)$ and $I_2(t)$ are the number of infectious individuals for the SIR and SEIR model, respectively. Interestingly, there are two peaks in the Jensen-Shannon divergence, towards earlier measurement times and around $t=60$. The prior predictive distributions are most similar where the means of the data distributions cross, i.e.~near $t=30$ (see the bottom left plot).

\subsection{Additional results}

We here present and discuss several additional results for the epidemiology experiments in the main text. 

Figure~\ref{fig:sde_pe_results} shows the training curves of the parameter estimation (PE) task for the SIR model with different experimental budgets $D$. For these experiments we maximised the JSD lower bound, which is shown in the top row as a function of training epochs. The elements of the experimental design vector are shown in the middle row and show a convergence towards early measurement times. The bottom row of Figure~\ref{fig:sde_pe_results} shows evaluations of the NWJ lower bound using the density ratio learned by maximising the JSD lower bound (as discussed in the main text). These NWJ lower bound evaluations show large fluctuations, in stark contrast to the smooth JSD lower bound curves. These fluctuations presumably arise because of the exponential term in the NWJ lower bound (see Equation 3.2 in the main text). Importantly, however, we only use the NWJ bound as a means to get an estimate of the mutual information, and we do not use it to update the neural network parameters or experimental designs. Thus, the training behaviour is not affected by these large fluctuations seen in the top row. Putting a cap on the exponential term in the NWJ lower bound may help reduce variance~\citep[as done in][]{Song2020} if that is required.

\begin{figure}[!t]
    \centering
    \includegraphics[width=\linewidth]{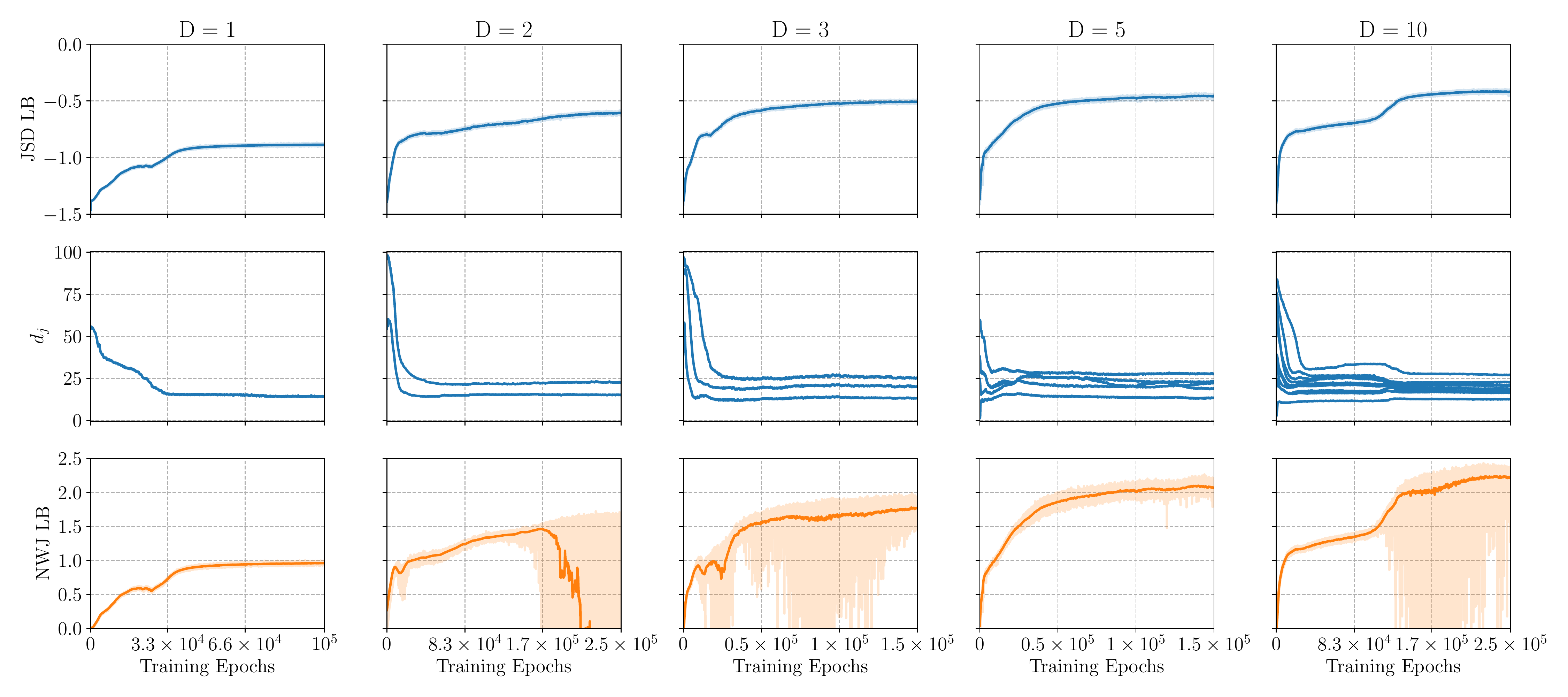}
    \vspace{-0.7cm}
    \caption[]{PE training curves for the SDE-based SIR model, with different experimental budgets $D$. Shown are JSD lower bound values (top row), the elements of the experimental design vector (middle row) and NWJ lower bound evaluations (bottom row). Importantly, the JSD lower bound was used to update neural network parameters and designs, while the NWJ lower bound was purely used to obtain MI estimates.}
    \label{fig:sde_pe_results}
\end{figure}

\begin{figure}[!t]
    \centering
    \includegraphics[width=\linewidth]{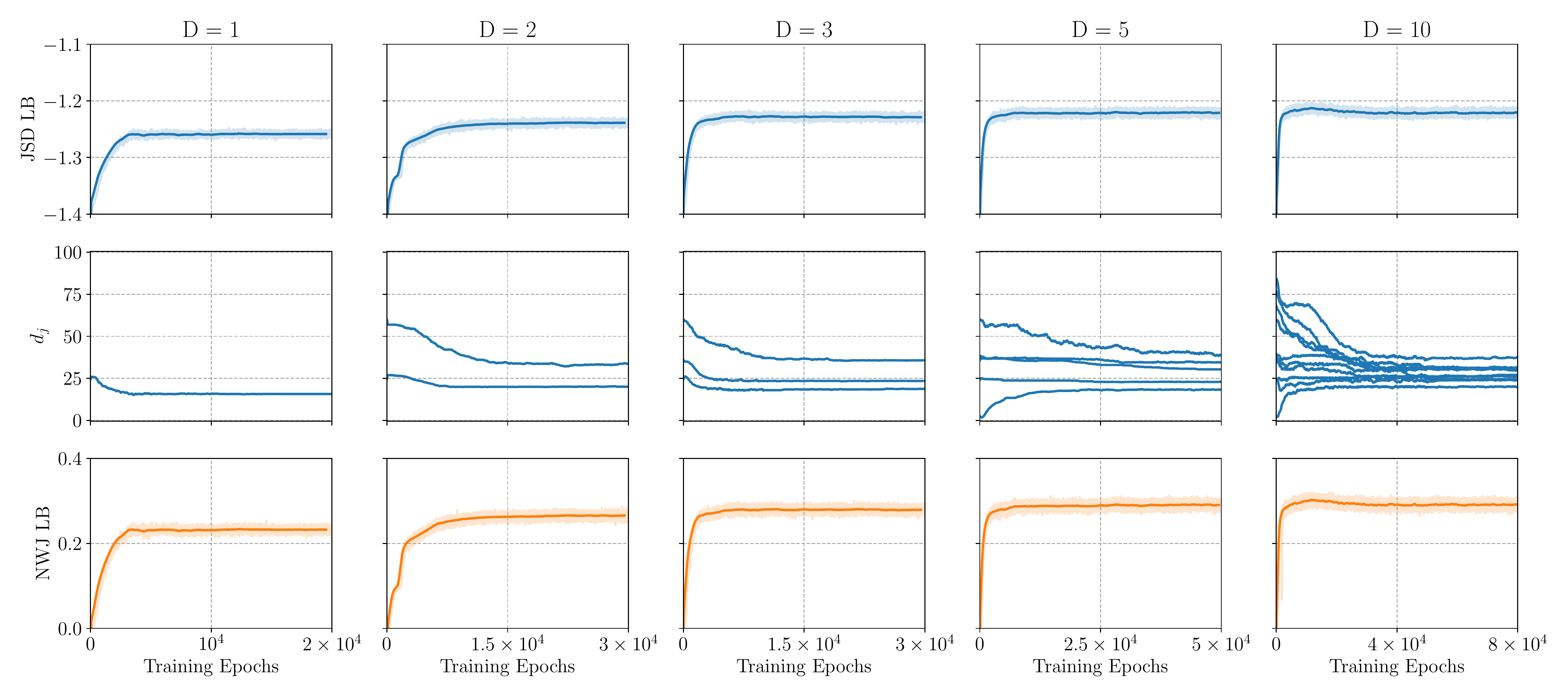}
    \vspace{-0.7cm}
    \caption[]{MD training curves for the SDE-based SIR and SEIR models, with different experimental budgets $D$. Shown are JSD lower bound values (top row), the elements of the experimental design vector (middle row) and NWJ lower bound evaluations (bottom row). Importantly, the JSD lower bound was used to update neural network parameters and designs, while the NWJ lower bound was purely used to obtain MI estimates.}
    \label{fig:sde_md_results}
\end{figure}

\begin{figure}[!t]
    \centering
    \includegraphics[width=\linewidth]{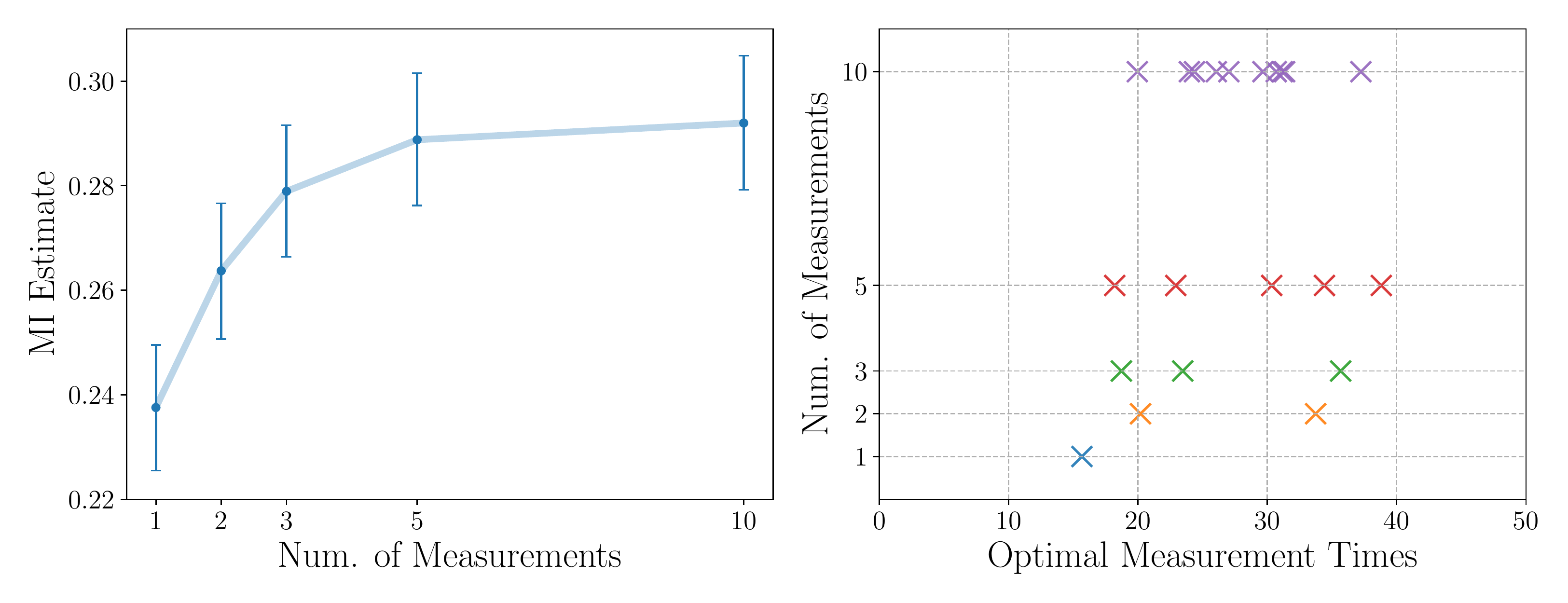}
    \vspace{-0.7cm}
    \caption[]{MD results for the SDE-based SIR and SEIR models with different experimental budgets. The left plot shows validation MI estimates for different number of measurements, averaged over several validation data sets. The right plot shows the corresponding optimal measurement times.}
    \label{fig:sde_md_mi}
\end{figure}

\begin{figure}[!h]
    \centering
    \includegraphics[width=\linewidth]{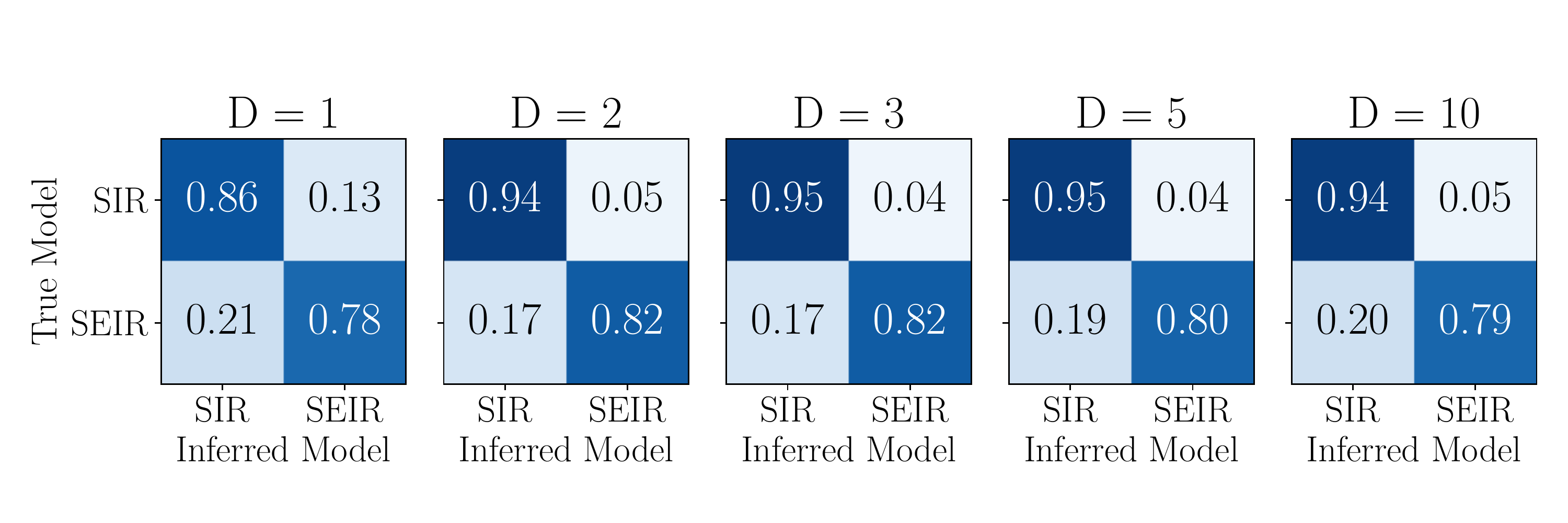}
    \vspace{-0.7cm}
    \caption[]{MD posteriors for the epidemiology models with different experimental budgets $D$. Shown are average posterior probabilities with different ground truths.}
    \label{fig:sde_md_posteriors}
\end{figure}

Similarly, Figure~\ref{fig:sde_md_results} shows the training curves of the model discrimination (MD) task for the SDE-based SIR and SEIR models, with different experimental budgets $D$, i.e.~number of measurements. Unlike for the PE task, the NWJ lower bound evaluations for the MD task do not have large fluctuations and are quite smooth. Validation estimates of the mutual information for different number of measurements are shown in Figure~\ref{fig:sde_md_mi}, as well as corresponding optimal designs (that are also shown in the main text). 
We show corresponding average posterior distributions for different ground truth models in Figure~\ref{fig:sde_md_posteriors}. Model recovery tends to improve as the number of measurements increases. However, the model recovery of the SEIR model, i.e.~the average posterior probability $\mathbb{E}[\widehat{p}(m=2|\ybf^\ast, \dbf^\ast, m_{\text{truth}}=2)]$, is always worse than that of the SIR model, i.e.~the average posterior probability $\mathbb{E}[\widehat{p}(m=1|\ybf^\ast, \dbf^\ast, m_{\text{truth}}=1)]$, which are shown by the diagonal entries in Figure~\ref{fig:sde_md_posteriors}. This may be because the optimal designs (shown in Figure~\ref{fig:sde_md_mi}) are all clustered towards earlier measurement times, i.e.~below $t=40$. Looking at the bottom left of Figure~\ref{fig:sde_data}, this is the region where the SIR model responses dominate over the SEIR model responses.

\clearpage
\bibliographystyle{bib/ba}
\bibliography{bib/references}

\paragraph{Acknowledgements}\mbox{}\\[1em]
Steven Kleinegesse was supported in part by the EPSRC Centre for Doctoral Training in Data Science, funded by the UK Engineering and Physical Sciences Research Council (grant EP/L016427/1) and the University of Edinburgh.

\end{document}